\documentclass[journal]{IEEEtran}  
\usepackage{amsmath,amsfonts}
\usepackage{algorithmic}
\usepackage{algorithm}
\usepackage{ragged2e} 
\usepackage{booktabs,makecell, multirow, tabularx}
\usepackage{array}
\usepackage[caption=false,font=normalsize,labelfont=sf,textfont=sf]{subfig}
\usepackage{textcomp}
\usepackage{stfloats}
\usepackage{url}
\usepackage{verbatim}
\usepackage{graphicx}
\usepackage{cite}
\usepackage{xcolor}
\begin{document}

\title{Give Us the Facts: Enhancing Large Language Models with Knowledge Graphs for Fact-aware Language Modeling}

\author{Linyao Yang, Hongyang Chen,~\IEEEmembership{Senior Member,~IEEE}, Zhao Li, Xiao Ding, Xindong Wu,~\IEEEmembership{Fellow,~IEEE}
\thanks{This work was supported in part by National Natural
Science Foundation of China under Grant 62306288, 62271452, National Key Research and Development Program of China (2022YFB4500305) and Key Research Project of Zhejiang Lab (No. 2022PI0AC01). 
 (Corresponding author: Hongyang Chen)}
\thanks{Linyao Yang, Hongyang Chen, Zhao Li, and Xindong Wu are with Zhejiang Lab, Hangzhou 311121, China (email: yangly@zhejianglab.com; dr.h.chen@ieee.org; zhaoli@zhejianglab.com; xwu@hfut.edu.cn)}
\thanks{Xiao Ding is with the Research Center for Social Computing and Information Retrieval, Harbin Institute of Technology, Harbin 150001, China (email: xding@ir.hit.edu.cn)}
}

\markboth{Journal of \LaTeX\ Class Files,~Vol.~14, No.~8, August~2021}%
{Shell \MakeLowercase{\textit{et al.}}: A Sample Article Using IEEEtran.cls for IEEE Journals}


\maketitle

\begin{abstract}
Recently, ChatGPT, a representative large language model (LLM), has gained considerable attention. Due to their powerful emergent abilities, recent LLMs are considered as a possible alternative to structured knowledge bases like knowledge graphs (KGs). However, while LLMs are proficient at learning probabilistic language patterns and engaging in conversations with humans, they, like previous smaller pre-trained language models (PLMs), still have difficulty in recalling facts while generating knowledge-grounded contents. To overcome these limitations, researchers have proposed enhancing data-driven PLMs with knowledge-based KGs to incorporate explicit factual knowledge into PLMs, thus improving their performance in generating texts requiring factual knowledge and providing more informed responses to user queries. This paper reviews the studies on enhancing PLMs with KGs, detailing existing knowledge graph enhanced pre-trained language models (KGPLMs) as well as their applications. Inspired by existing studies on KGPLM, this paper proposes enhancing LLMs with KGs by developing knowledge graph-enhanced large language models (KGLLMs). KGLLM provides a solution to enhance LLMs' factual reasoning ability, opening up new avenues for LLM research.


\end{abstract}


\section{Introduction}
\IEEEPARstart{I}{n} recent years, the rapid development of big data \cite{9094012,9720100,8444740} and high-speed computing has led to the emergence of pre-trained language models (PLMs). Plenty of PLMs, such as BERT \cite{BERT}, GPT \cite{GPT-1}, and T5 \cite{T5}, have been proposed, which greatly improve the performance of various natural language processing (NLP) tasks. Recently, researchers have found that scaling model size or data size can improve model capacities on downstream tasks. Moreover, they found that when the parameter size exceeds a certain scale \cite{wei2022emergent}, these PLMs exhibit some surprising emergent abilities. Emergent abilities refer to the abilities that are not present in small models but arise in large models \cite{wei2022emergent}, which are utilized to distinguish large language models (LLMs) from PLMs. 


On November 30, 2022, a chatbot program named ChatGPT was released by OpenAI, which is developed based on the LLM GPT-3.5. By fine-tuning GPT with supervised learning and further optimizing the model using reinforcement learning from human feedback (RLHF), ChatGPT is capable of engaging in continuous conversation with humans based on chat context. It can even complete complex tasks such as coding and paper writing, showcasing its powerful emergent abilities \cite{wei2022emergent}. Consequently, some researchers \cite{LAMA,MAMA,heinzerling-inui-2021-language,bian2023chatgpt} explored whether LLMs can serve as parameterized knowledge bases to replace structured knowledge bases like knowledge graphs (KGs), as they also store a substantial amount of facts.


However, existing studies \cite{wang2021can,cao-etal-2021-knowledgeable,liu2023evaluating,bang2023multitask} have found that LLMs' ability to generate factually correct text is still limited. They are capable of remembering facts only during training. Consequently, these models often face challenges when attempting to recall relevant knowledge and apply the correct knowledge to generate knowledge grounded contents. On the other hand, as artificially constructed structured knowledge bases, KGs store a vast amount of knowledge closely related to real-world facts in a readable format. They explicitly express relationships between entities and intuitively display the overall structure of knowledge and reasoning chains, making them an ideal choice for knowledge modeling. As a result, there exists not only a competitive but also a complementary relationship between LLMs and KGs. LLMs have the ability to enhance knowledge extraction accuracy and improve the quality of KGs \cite{3507066}, while KGs can utilize explicit knowledge to guide the training of LLMs, improving their ability to recall and apply knowledge.


So far, numerous methods have been proposed for strengthening PLMs with KGs, which can be categorized into three types: before-training enhancement, during-training enhancement, and post-training enhancement. Although there exist a few surveys \cite{wei2021knowledge,yang2021survey,zhen2022survey} of knowledge-enhanced PLMs, they focus on various forms of knowledge, lacking a systematic review of knowledge graph enhanced pre-trained language model (KGPLM) methods. For instance, Wei \textit{et al.} \cite{wei2021knowledge} conducted a review of knowledge enhanced PLMs based on diverse knowledge sources but only covered a small set of KGPLMs. Similarly, Yang \textit{et al.} \cite{yang2021survey} covered various forms of knowledge enhanced PLMs but provided only a partial review of KGPLMs without technical categorization. In another study, Zhen \textit{et al.} \cite{zhen2022survey} categorized knowledge enhanced PLMs into implicit incorporation and explicit incorporation methods, yet their review encompassed only a small subset of KGPLMs. Moreover, this field is rapidly evolving with numerous new technologies consistently being introduced. Therefore, to address questions of whether constructing KGs is still necessary and how to improve the knowledge modeling ability of LLMs, we present a systematic review of relevant studies. We conducted a thorough search for papers related to the keywords "language model" and "knowledge graph". Subsequently, the papers that were most relevant to KGPLM were carefully refined and categorized. In comparison with existing surveys, this paper specifically concentrates on KGPLM and covers a broader range of up-to-date papers. Furthermore, we suggest the development of knowledge graph enhanced large language models (KGLLMs) to tackle the knowledge modeling challenge in LLMs. The main contributions of this paper are summarized as follows:

\begin{itemize}
    \item We provide a comprehensive review for KGPLMs, which helps researchers to gain a deep insight of this field.
    \item We overview research on the evaluation of LLMs and draw comparisons between LLMs and KGs. 
    \item We propose to enhance LLMs with KGs and suggest some possible future research directions, which may benefit researchers in the field of LLM.
\end{itemize}

The remainder of this paper is organized as follows. Section \uppercase\expandafter{\romannumeral2} overviews the background of LLMs. Section \uppercase\expandafter{\romannumeral3} categorizes the existing methods for KGPLMs and introduces representatives from each group. Section \uppercase\expandafter{\romannumeral4} introduces the applications of KGPLMs. Section \uppercase\expandafter{\romannumeral5} discusses whether LLMs can replace KGs with the evidence from existing studies. Section \uppercase\expandafter{\romannumeral6} proposes to enhance LLMs' ability to learn factual knowledge by developing KGLLMs and presents some future research directions. Section \uppercase\expandafter{\romannumeral7} draws the conclusions.

\section{Background}

PLMs learn dense and continuous representations for words, addressing the issue of feature sparsity encountered in traditional encoding methods and significantly improving performance across various NLP tasks. Consequently, PLM-based methods have gained prominence, leading to the development of various types of PLMs. Recently, PLMs have been scaled to LLMs in order to achieve even better performance. In this section, we provide a comprehensive background of PLMs and offer an overview of their historical development.

\begin{figure*}
    \centering
    \includegraphics[width=0.8\textwidth]{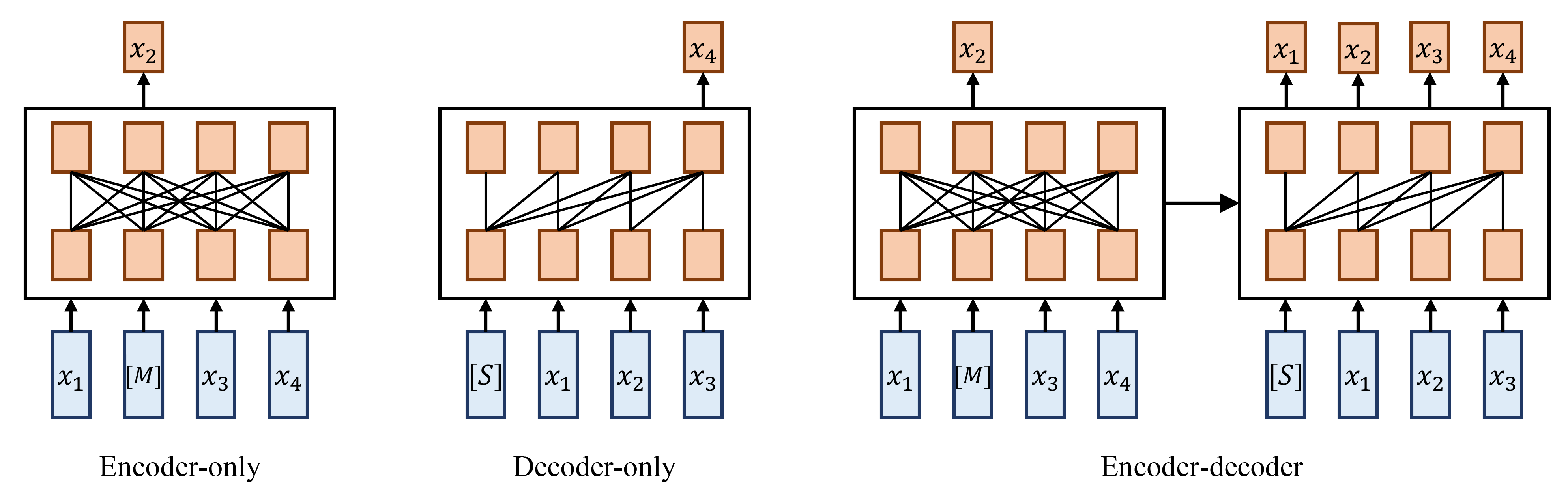}
    \caption{Main frameworks of existing PLMs, in which $x_{i}$ is the $i$-th token of the input sentence, $\left [ M \right ]$ represents the masked token and $\left [ S \right ]$ is the start token.}
    \label{fig:framework}
\end{figure*}

\subsection{Background of PLMs}

PLMs are a type of language model obtained through unsupervised learning \cite{9462394} on a large corpus. They are capable of capturing the structure and characteristics of a language and generating universal representations for words. Following pre-training, PLMs can be fine-tuned for specific downstream tasks like text summarization, text classification, and text generation.

The model frameworks used by existing PLMs can be classified into three categories, as illustrated in Fig.~\ref{fig:framework}: encoder-only, decoder-only, and encoder-decoder \cite{wang2022pre}. The encoder-only framework utilizes a bidirectional transformer to recover masked tokens based on the input sentences, which effectively utilizes contextual information to learn better text representations. More specifically, given an input token sequence $\mathcal{\not{C}} = \left (x_{1},...,x_{T} \right )$ with a few masked tokens $\mathcal{M}$, it models the likelihood of the masked tokens as $p(x)=\sum_{x_{t}\in \mathcal{M}}p(x_{t}|x_{\mathcal{\not{C}}  }  )$. However, due to the lack of a decoder, it cannot be directly applied to text generation tasks. BERT and its improved models mostly adopt the encoder-only framework. The decoder-only framework leverages a unidirectional transformer to predict tokens in an autoregressive fashion, making it suitable for text generation tasks. That is, given the text sequence $\mathcal{C} =\left ( x_{1},...,x_{T} \right )$, this framework models the likelihood of the input token sequence as $p(x)=\prod_{t=1}^{T}p(x_{t}|x_{<t})$. GPT series and their improved models mostly adopt this framework. Nevertheless, compared with the other two frameworks, the decoder-only framework cannot make use of contextual information and cannot generalize well to other tasks. The encoder-decoder framework constructs a sequence-to-sequence model to predict the current token based on historical context with masked tokens. Its objective can be described as $\sum_{t=1}^{T}p(x_{t}|x_{<t,\mathcal{\not{C}}  } ) $. This framework excels at tasks that require generating output based on given inputs, yet its encoding and decoding speed is slow compared to the other two frameworks.


\begin{figure*}
    \centering
    \includegraphics[width=0.88\textwidth]{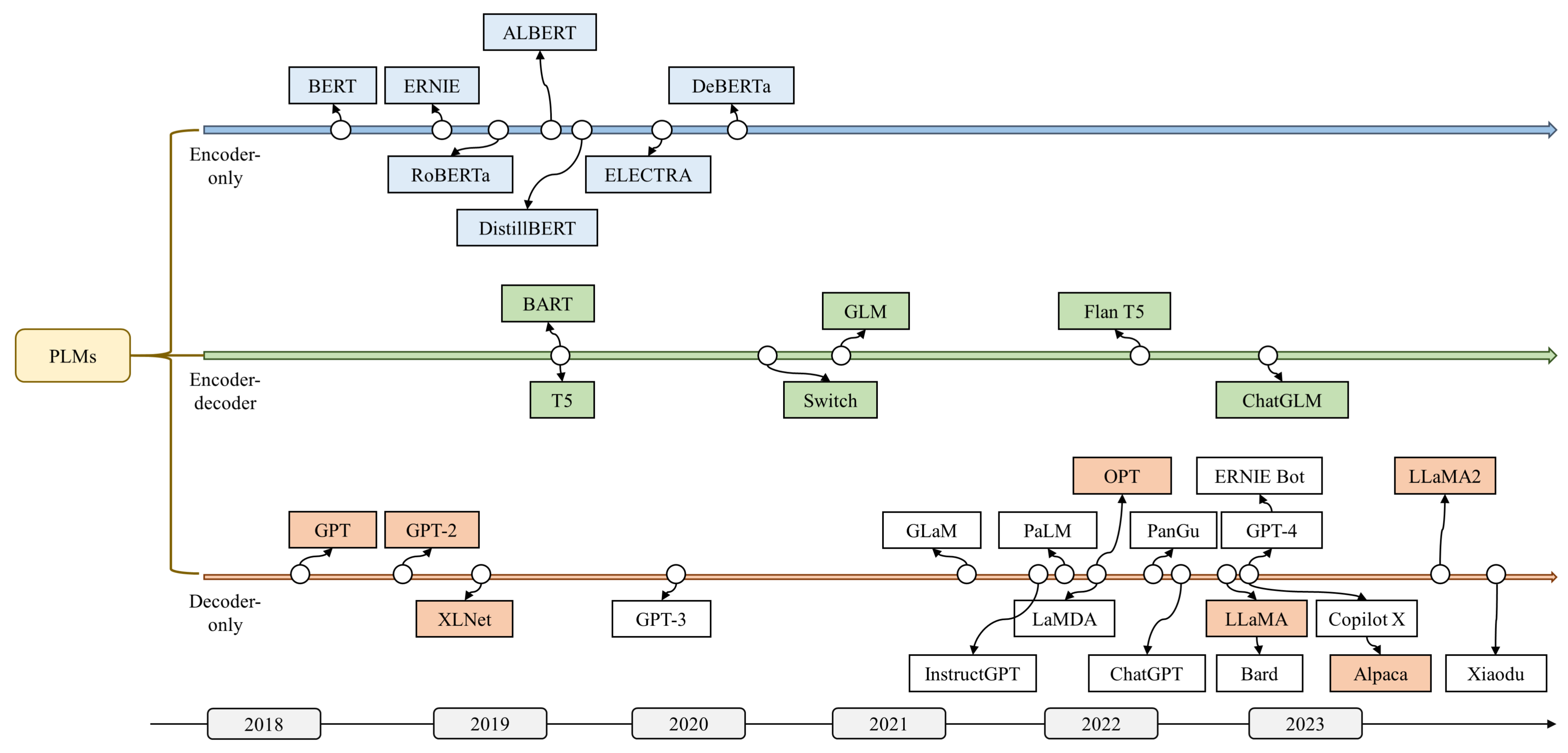}
    \caption{Milestones of LLMs. Open-source models are represented by solid squares, while closed-source models are represented by hollow squares.}
    \label{fig:milestone}
\end{figure*}

Multiple pre-training tasks for PLMs have been designed, which can be categorized into word-level, phrase-level, and sentence-level tasks. Typical word-level pre-training tasks include masked language modeling (MLM) \cite{BERT} and replaced token detection (RTD) \cite{Electra}. MLM randomly masks some tokens in the input sequence and trains PLMs to reconstruct the masked tokens based on context, whose loss function is:
\begin{equation}
    \mathcal{L}_{\mathrm{MLM} }=-\sum_{x\in \mathcal{M}}\log{p}(x|x_{\not{\mathcal{C}} } ).
\end{equation}
It can promote the learning of contextual information, thereby achieving better results in language understanding and language modeling tasks. RTD operates similarly to MLM but introduces greater randomness by substituting some tokens with alternative ones and training the model to predict the original tokens, whose loss function is defined as:
\begin{equation}
    \mathcal{L}_{\mathrm{RTD} }=-\sum_{t=1}^{T}\log{p(y_{t}|\tilde{x} )}.
\end{equation}
Here, $\tilde{x}$ is the corrupted token of $x$, while $y_{t}$ is 1 if $\tilde{x}_{t}=x_{t} $ and 0 otherwise. Compared with MLM, RTD can reflect changes in vocabulary in real texts more realistically and enable PLMs to handle unknown and misspelled words. The representative of phrase-level pre-training tasks is span boundary objective (SBO) \cite{SpanBERT,SPE}, which forces PLMs to predict each token of a masked span solely relying on the representations of the visible tokens at the boundaries, enhancing the syntactic structure analysis ability of PLMs and improving their performance in named entity recognition and sentiment analysis. The training objective of the SBO task can be expressed as:
\begin{equation}
    \mathcal{L}_{\mathrm{SBO} } = -\sum_{t=1}^{T}\log{p(x_{i}|y_{i})},
\end{equation}
where $y_{i}$ is token $x_{i}$'s representation in the span. Representatives of sentence-level pre-training tasks include next sentence prediction (NSP) \cite{BERT} and sentence order prediction (SOP) \cite{ALBERT}. NSP trains PLMs to distinguish whether two given sentences are continuous, thereby improving PLMs' performance in context-based tasks such as natural language inference and text classification. Similarly, SOP trains PLMs to determine the order of two randomly sampled and disrupted sentences, which improves their ability to capture sentence order information. The training objective of NSP and SOP is as follows:
\begin{equation}
    \mathcal{L}_{\mathrm{NSP}/SOP} = -\log{p(y|s_{1},s_{2})},
\end{equation}
where $y=1$ if $s_{1}$ and $s_{2}$ are two consecutive segments extracted from the corpus. Other tasks like deleted token detection (DTD), text infilling, sentence reordering (SR), and document reordering (DR) are also utilized by some PLMs \cite{BART}, which improve their performance in some special tasks.

\begin{figure*}
    \centering
    \includegraphics[width=0.9\textwidth]{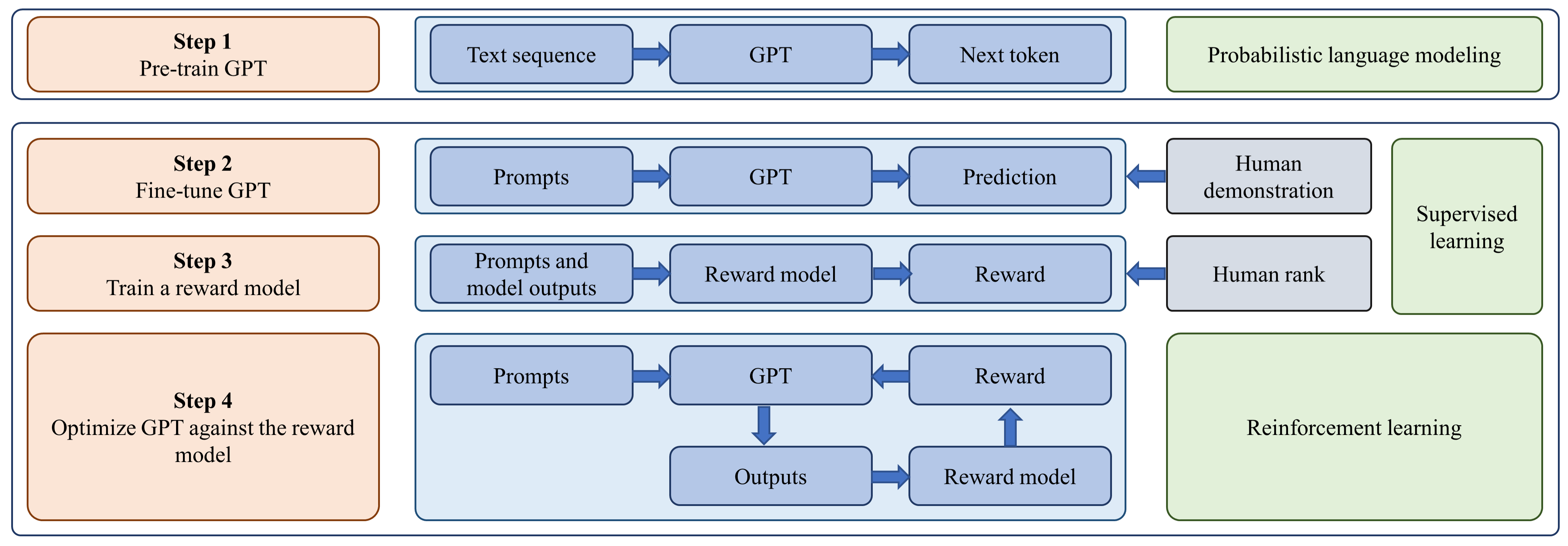}
    \caption{The implementation process of ChatGPT.}
    \label{fig:ChatGPT}
\end{figure*}


\subsection{Milestones}

As an early attempt, Elmo \cite{Elmo} employs a bidirectional long short term memory (LSTM) network to learn word representations capturing context. The model is trained with a bidirectional autoregressive language modeling objective, which involves maximizing the following log-likelihood:
\begin{equation}
\begin{aligned}
& \sum_{k=1}^T\left(\log p\left(x_t \mid x_1, \ldots, x_{t-1} ; \Theta_x, \overrightarrow{\Theta}_{L S T M} \right)\right. \\
& \left.\quad+\log p\left(x_t \mid x_{t+1}, \ldots, x_T ; \Theta_x, \overleftarrow{\Theta}_{L S T M} \right)\right),
\end{aligned}
\end{equation}
where $p$ models the probability of token $x_{t}$ given the history context $\left(x_1, \ldots, x_{t-1}\right)$ or the future context $\left(x_{t+1}, \ldots, x_T\right)$. $\Theta_x$ denotes the token representation. $\overrightarrow{\Theta}_{L S T M}$ and $\overleftarrow{\Theta}_{L S T M}$ denote the LSTM encoder in the forward direction and the backward direction, respectively. By learning context-aware word representations, Elmo largely raises the performance bar of NLP tasks. However, its feature extraction ability is limited since LSTM is difficult to handle long sequences. With the emergence of the highly parallelizable Transformer \cite{Transformer}, more powerful contextualized PLMs have been developed. Notable PLMs with different framewroks are shown in Fig.~\ref{fig:milestone}.

Transformer employs a self-attention mechanism to capture the dependence among input sequences, allowing for parallel processing of tokens and improving efficiency. Specifically, the output from the self-attention mechanism is:
\begin{equation}
    \mathbf{h} = softmax(\frac{\mathbf{QK} ^{T}}{\sqrt{d_{k} } } )\mathbf{V},
\end{equation}
where $\mathbf{Q}$, $\mathbf{K}$, and $\mathbf{V}$ are the query matrix, key matrix, and value matrix. $d_{k}$ is the dimension of the key and query vectors. 

Encoder-only PLMs utilize bidirectional Transformer as encoder and employ MLM and NSP tasks for self-supervised training. RoBERTa \cite{RoBERTa} introduces a set of design choices and training strategies that lead to better performance, significantly enhancing BERT's performance on various benchmarks. DistilBERT \cite{DistilBERT} incorporates knowledge distillation into pre-training, which reduces the size of a BERT model by 40\%. Other notable encoder-only PLMs include ERNIE \cite{ERNIE}, ALBERT \cite{ALBERT}, ELECTRA \cite{Electra}, and DeBERTa \cite{DeBERTa}.

In contrast, in decoder-only PLMs, a unidirectional Transformer is utilized as decoder, and the model is trained to predict the next token based on the preceding sequence. This training approach improves their language understanding and text generation abilities. Given an unsupervised corpus, GPT uses a unidirectional language modeling objective to optimize the model, maximizing the following log-likelihood:
\begin{equation}
    \sum_{i}logp(x_{t}|x_{t-k},...,x_{t-1};\Theta ).
\end{equation}
Here, $\Theta$ represents the parameters of the Transformer model. GPT-2 \cite{GPT-2} improves upon GPT by increasing its model size and training corpus and enabling the model to automatically recognize task types for unsupervised training. XLNet \cite{XLNet} proposes a generalized autoregressive pretraining method, which enables learning bidirectional contexts.

In encoder-decoder PLMs, Transformer serves as both encoder and decoder. The encoder generates the latent representations for the input sequence, while the decoder generates the target output text. T5 \cite{T5} develops a unified framework that converts all NLP tasks into a text-to-text format, leading to exceptional performance on numerous benchmarks. In order to efficiently pre-train sequence-to-sequence models, BART \cite{BART} adopts a standard neural machine translation architecture and develops a denoising autoencoder.

\begin{table*}[!ht]
    \setlength\tabcolsep{7pt}
    \caption{Comparison of different PLMs}
    \centering
    \begin{tabular}{ccccccc}
    \hline
        Model framework & PLM & Year & Base model & Pre-training tasks & Pre-training data size & model size \\ \hline
        \multirow{7}{*}{Encoder-only} & BERT & 2018 & Transformer & MLM, NSP & 3300M words & 340M \\ 
        ~ & ERNIE & 2019 & Transformer & MLM, NSP & 4500M subwords & 114M \\ 
        ~ & RoBERTa & 2019 & BERT & MLM & 160GB of text & 335M \\ 
        ~ & ALBERT & 2019 & BERT &  SOP & 16GB of text & 233M \\ 
        ~ & DistillBERT & 2019 & BERT &  MLM & 3300M words & 66M \\ 
        ~ & ELECTRA & 2020 &  Transformer &  RTD & 126GB of text & 110M \\ 
        ~ & DeBERTa & 2020 & Transformer & MLM & 78GB of text & 1.5B \\ \hline
        \multirow{6}{*}{Encoder-decoder} & BART & 2019 & Transformer & MLM, DTD, text infilling, SR, DR & 160GB of text & 406M \\ 
        ~ & T5 & 2019 & Transformer & MLM & 20TB of text & 11B \\ 
        ~ & Switch & 2021 & Transformer & MLM & 180B tokens & 1.6T \\ 
        ~ & GLM & 2021 & Transformer & Blank infilling & 400B tokens & 130B \\
        ~ & Flan T5 & 2022 & T5 & 1800 fine-tuning tasks & - & 11B \\ 
        ~ & ChatGLM & 2023 & GLM & Blank infilling & 1T tokens & 6B \\ \hline
        \multirow{14}{*}{Decoder-only} & GPT & 2018 & Transformer & Autoregressive language modeling & 800M words & 117M \\ 
        ~ & GPT-2 & 2019 & Transformer & Autoregressive language modeling & 40GB of text & 1.5B \\ 
        ~ & XLNet & 2019 & Transformer & Autoregressive language modeling & 33B tokens & 340M \\ 
        ~ & GPT-3 & 2020 & Transformer & Autoregressive language modeling & 45TB of text & 175B \\ 
        ~ & GLaM & 2021 & Transformer & Autoregressive language modeling & 1.6T tokens & 1.2T \\ 
        ~ & InstructGPT & 2022 & GPT-3 & Autoregressive language modeling & - & 175B \\ 
        ~ & PaLM & 2022 & Transformer & Autoregressive language modeling & 780B tokens & 540B \\ 
        ~ & LaMDA & 2022 & Transformer & Autoregressive language modeling & 768B tokens & 137B \\ 
        ~ & OPT & 2022 & Transformer & Autoregressive language modeling & 180B tokens & 175B \\ 
        ~ & ChatGPT & 2022 & GPT-3.5 & Autoregressive language modeling & - & - \\ 
        ~ & LLaMA & 2023 & Transformer & Autoregressive language modeling & 1.4T tokens & 65B \\ 
        ~ & GPT-4 & 2023 & Transformer & Autoregressive language modeling & 13T tokens & 1.8T \\ 
        ~ & Alpaca & 2023 & LLaMA & Autoregressive language modeling & 52K data & 7B \\ 
        ~ & LLaMA2 & 2023 & Transformer & Autoregressive language modeling & 2T tokens & 70B \\ \hline
    \end{tabular}
    \label{tab:plm}
\end{table*}

\subsection{Scaling PLMs to LLMs}

With the emergence of more and more PLMs, it has been revealed that model scaling can lead to improved performance. By increasing the parameter scale and data scale to a large enough size, it was found that these enlarged models exhibit some special abilities that do not possess by small-scale PLMs. Therefore, recent efforts have been devoted to scaling PLMs to LLMs to empower them with emergent abilities. Typically, LLMs refer to PLMs that consist of hundreds of billions of parameters, such as GLM \cite{GLM}, Switch \cite{Switch}, Flan T5 \cite{FlanT5}, and ChatGLM \cite{ChatGLM} of the encoder-decoder framework. Besides, most existing LLMs adopt the decoder-only framework. Notable examples of decoder-only LLMs include  GPT-3 \cite{GPT-3}, GLaM \cite{GLaM}, InstructGPT \cite{InstructGPT}, PaLM \cite{PaLM-base}, LaMDA \cite{Lamda}, OPT \cite{OPT}, LLaMA \cite{LLaMA}, Alpaca \cite{Alpaca}, GPT-4 \cite{GPT-4}, and LLaMA2 \cite{LLaMA2}. GPT-3 \cite{GPT-3} further increases GPT-2's parameters and its training data size, and adopts zero-shot learning and diversity generation technologies, making it possible to learn and execute new tasks without annotated data and generate texts with diverse styles. GPT-3.5 not only increases the model size but also applies novel pre-training methods such as prompt-based extraction of templates (PET), which further improves the accuracy and fluency of generated texts. LLMs have stronger abilities to understand natural language and solve complex NLP tasks than smaller PLMs. GPT-3, for instance, exhibits a remarkable in-context learning ability. It can generate expected outputs for test cases by filling in the word sequence of the input text, relying solely on natural language instructions or demonstrations, without the need for additional training. Conversely, GPT-2 lacks this ability \cite{zhao2023survey}.

The most remarkable application of LLMs is ChatGPT, which adapts GPT-3.5 for dialogue and demonstrates an amazing conversation ability. The implementation process of ChatGPT is shown in Fig.~\ref{fig:ChatGPT} \cite{Lujingwei}. It first trains GPT on a large-scale corpus and then fine-tunes it on a dataset of labeler demonstrations. After that, it optimizes the model using RLHF \cite{RLHF}, which trains a reward model to learn from direct feedback provided by human evaluators and optimizes the GPT model by formulating it as a reinforcement learning problem. In this setting, the pre-trained GPT model serves as the policy model that takes small pieces of prompts \cite{Prompt-learning} as inputs and returns output texts. The GPT policy model is then optimized using the proximal policy optimization (PPO) algorithm \cite{PPO} against the reward model. Based on the RLHF method, ChatGPT enables GPT to follow the expected instructions of humans and reduces the generation of toxic, biased, and harmful content. Besides, ChatGPT adopts the chain-of-thought strategy \cite{wei2022chain} and is additionally trained on code data, enabling it to solve tasks that require intermediate logical steps.


Another notable advancement is GPT-4 \cite{GPT-4}, a model that extends text input to multimodal signals and exhibits greater proficiency at solving tasks \cite{bubeck2023sparks}. Furthermore, GPT-4 has undergone six months of iterative alignment, adding an additional safety reward in the RLHF training, which has made it more adept at generating helpful, honest, and harmless content. Additionally, GPT-4 implements some enhanced optimization methods, such as predictable scaling that accurately predicts GPT-4's final performance from smaller models trained with less computation.

Table~\ref{tab:plm} summarizes the characteristics of the above context-based PLMs and LLMs. As observed, the parameter size of the largest model has increased year by year. 

\subsection{Pros and Cons of LLMs}


A proliferation of benchmarks and tasks has been leveraged to evaluate the effectiveness and superiority of LLMs. Results from corresponding experiments demonstrate that LLMs achieve much better performance than previous deep learning models and smaller PLMs on a variety of NLP tasks. Besides, LLMs exhibit some emergent abilities and are capable of solving some complex tasks that traditional models and smaller PLMs cannot address. In summary, LLMs have the following superior characteristics.

\textbf{Zero-shot Learning.} LLMs outperform other models with zero-shot learning on most tasks and even perform better than fine-tuned models on some tasks. An empirical study  \cite{bang2023multitask} has shown that ChatGPT outperforms previous models with zero-shot learning on 9 of 13 datasets and even outperforms fully fine-tuned task-specific models on 4 tasks. This superior performance is attributed to the rich and diverse input data as well as the large parameter scale of LLMs, which allow them to capture the underlying patterns of natural language with high fidelity, leading to more robust and accurate inferences.

\textbf{In-context Learning.} In-context learning (ICL) is a paradigm that allows LLMs to learn tasks from only a few instances in the form of demonstration \cite{ICL}. ICL was exhibited for the first time by GPT-3, which has become a common approach to use LLMs. ICL employs a formatted natural language prompt, which includes a description of the task and a handful of examples to illustrate the way to accomplish it. The ICL ability also benefits from the strong sequence processing ability and the rich knowledge reserve of LLMs.

\textbf{Step-by-step Reasoning.} By utilizing the chain-of-thought prompting strategy, LLMs can successfully complete some complex tasks, including arithmetic reasoning, commonsense reasoning, and symbolic reasoning. Such tasks are typically beyond the capability of smaller PLMs. The chain-of-thought is an improved prompting strategy, which integrates intermediate reasoning steps into the prompts to boost the performance of LLMs on complex reasoning tasks. Besides, the step-by-step reasoning ability is believed to be potentially acquired through training LLMs on well-structured code data \cite{wei2022chain}.

\textbf{Instruction Following.} Instruction tuning is a unique fine-tuning approach that fine-tunes LLMs on a collection of natural language formatted instances. With this approach, LLMs are enabled to perform well on previously unseen tasks described through natural language instructions without relying on explicit examples \cite{zhao2023survey}. For example, Wei \textit{et al.} \cite{wei2021finetuned} fine-tuned a 137B parameter LLM on over 60 datasets based on instruction tuning and tested it on unseen task types. The experimental results demonstrated that the instruction-tuned model significantly outperformed its unmodified counterpart and zero-shot GPT-3.

\textbf{Human Alignment.} LLMs can be trained to generate high-quality, harmless responses that align with human values through the technique of RLHF, which involves incorporating humans into the training loop using carefully designed labeling strategies. RLHF comprises three steps: 1) collecting a labeled dataset consisting of input prompts and target outputs to fine-tune LLMs in a supervised way; 2) training a reward model on the assembled data, and 3) optimizing LLMs by formulating its optimization as a reinforcement learning problem. With this approach, LLMs are enabled to generate appropriate outputs that adhere to human expectations.

\textbf{Tools Manipulation.} Traditional PLMs are trained on plain text data, which limits their ability to solve non-textual tasks. Besides, their abilities are limited by the pre-training corpus, and cannot effectively solve tasks requiring real-time knowledge. In response to these limitations, recent LLMs are developed with the ability to manipulate external tools such as search engine, calculator, and compiler to enhance their performance in specialized domains \cite{qin2023tool}. More recently, the plugin mechanism has been supported in LLMs, providing an avenue for implementing novel functions. This mechanism has significantly broadened the range of capacities for LLMs, making them more flexible and adaptable to diverse tasks.

Although LLMs have made significant progress in natural language understanding and human-like content generation, they still have the following limitations and challenges \cite{zhao2023survey}.

\textbf{Unstructured Generation.} LLMs commonly rely on natural language prompts or instructions to generate text under specific conditions. This mechanism presents challenges for precisely constraining the generated outputs according to fine-grained or structural criteria. Ensuring specific text structures, such as the logical order of concepts throughout the entire text, can be difficult. This difficulty is amplified for tasks requiring formal rules or grammar. This is because LLMs mainly focus on the local context information of words and sentences during pre-training, while ignoring global syntactic and structural knowledge. A proposal for addressing this problem is to adopt an iterative prompting approach in generating text \cite{Re3}, mimicking the process of human writing. In contrast, KGs offer a structured summary and emphasize the correlation of relevant concepts when complex events involving the same entity extend across multiple sentences \cite{3512467}, thus enhancing the process of structured text generation.

\textbf{Hallucination.} When generating factual or knowledge-grounded texts, LLMs may produce content that contradicts existing sources or lack supporting evidence. This challenge widely occurs in existing LLMs and is known as the problem of hallucination, which results in a drop in their performance and poses risks when deploying them for real-world applications. The cause of this issue is related to LLMs' limited ability to utilize correct internal and external knowledge during task-solving. To alleviate this problem, existing studies have resorted to alignment tuning strategies, which incorporate human feedback to fine-tune LLMs. KGs provide structured and explicit representations of knowledge, which can be dynamically incorporated to augment LLMs, resulting in more factual rationales and reduced hallucination in generation \cite{li2023chain}.


\textbf{Inconsistency.} With the help of the chain-of-thought strategy, LLMs are capable of solving some complex reasoning tasks based on step-by-step reasoning. Despite their superior performance, LLMs may at times arrive at the desired answer based on an invalid reasoning path or produce an incorrect answer despite following a correct reasoning process. As a result, inconsistency arises between the derived answer and the underlying reasoning process. Additionally, research \cite{elazar2023measuring} has revealed that LLMs' abilities to forecast facts and answer queries are highly influenced by specific prompt templates and related entities. This is because that LLMs rely largely on simple heuristics to make predictions, their generations are correlated with co-occurrence frequencies between the target word and words in the prompt. Moreover, although LLMs' pre-training process helps them memorize facts, it fails to imbue them with the ability to generalize observed facts, leading to poor inferences. This issue can be partially addressed by introducing external KGs in LLM reasoning. By interactively exploring related entities and relations on KGs and performing reasoning based on the retrieved knowledge, LLMs can have better ability of knowledge traceability and knowledge correctability \cite{TOG}.


\textbf{Limited Reasoning Ability.} LLMs have demonstrated decent performance on some basic logical reasoning tasks when provided with question-answer examples. However, they exhibit poor performance on tasks that require the ability to comprehend and utilize supporting evidence for deriving conclusions. While LLMs typically generate valid reasoning steps, they face challenges when multiple candidate steps are deemed valid \cite{saparov2023language}. This results from LLMs being primed to solely choose the answer with the highest word overlapping with the input question. Additionally, LLMs struggle with predicting entity relationships due to their emphasis on shallow co-occurrence and sequence patterns of words. Moreover, despite exhibiting some basic numerical and symbolic reasoning abilities \cite{chang2023language}, LLMs face difficulties in numerical computation, especially for symbols infrequently encountered during pre-training. KGs explicitly capture the relations among concepts, which are essential for reasoning and can be utilized to enhance LLMs with structural reasoning capabilities. Previous studies have demonstrated that the integration of textual semantics and structural reasoning yields significant enhancement in the reasoning ability of LLMs \cite{wang2023unifying,GreaseLM}.

\begin{figure*}
    \centering
    \includegraphics[width=0.9\textwidth]{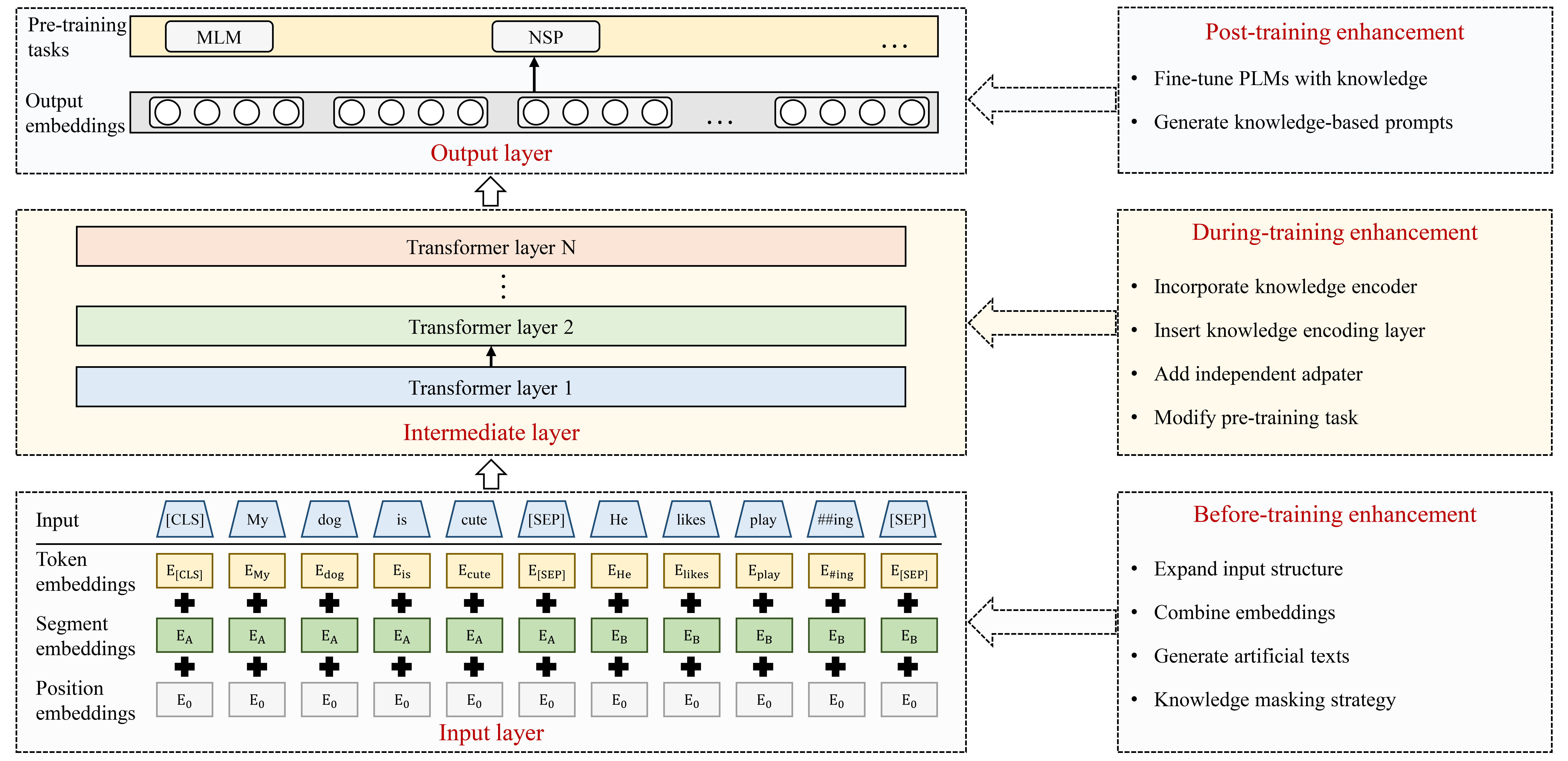}
    \caption{Three types of KGPLMs according to the stage of knowledge graph participating in pre-training.}
    \label{fig:kgplm}
\end{figure*}

\textbf{Insufficient Domain Knowledge.} Because of the limited availability of domain-specific corpus, LLMs may not perform as well on domain-specific tasks as on general ones. For instance, while such models generally capture frequent patterns from general texts, generating medical reports, which involve numerous technical terms, may pose a great challenge for LLMs. This limitation suggests that during pre-training, it is difficult for LLMs to acquire sufficient domain knowledge, and injecting additional specialized knowledge may come at the cost of losing previously learned information, given the issue of catastrophic forgetting. Therefore, developing effective techniques for knowledge injection is of critical importance to enhance the performance of LLMs on specialized domains. Domain KGs are effective and standardized knowledge bases for specific domains, offering a feasible source for unified domain knowledge. For example, Ding \textit{et al.} \cite{KnowledgeDA} proposed a unified domain LLM development service that leverages domain KGs to enhance the training process, which effectively improves LLMs' performance on domain-specific tasks.

\textbf{Knowledge Obsolescence.} LLMs are pre-trained on prior texts, thus limiting their ability to learn beyond the training corpus. This often results in poor performance when handling tasks that require most-recent knowledge. A simple solution to address this limitation is periodic retraining of LLMs on new data. However, the cost of such retraining is generally high. Hence, it is crucial to devise effective and efficient methods of incorporating current knowledge into LLMs. Prior studies have suggested using plugins as search engines for accessing up-to-date information. Nevertheless, these methods seem inadequate due to the difficulty of directly integrating specific knowledge into LLMs. Compared to LLMs, KGs offer a more straightforward update process that does not necessitate additional training. Updated knowledge can be incorporated into the input in the form of prompts, which are subsequently utilized by LLMs to generate accurate responses \cite{NLRSE}.

\textbf{Bias, Privacy, and Toxicity.} Although LLMs are trained to align with human expectations, they sometimes generate harmful, fully biased, offensive, and private content. When users interact with LLMs, models can be induced to generate such text, even without prior prompting or prompted with safe text. In fact, it has been observed that LLMs tend to degenerate into generating toxic text within just 25 generations \cite{Realtoxicityprompts}. Furthermore, despite their seemingly convincing text, LLMs generally tend to offer unhelpful and sometimes unsafe advice. For example, it has been revealed that GPT-3 produces worse advice than humans do in over 95\% of the situations described on Reddit \cite{Turingadvice}. The reasons are that such biased, private, and toxic texts widely exist in the pre-training corpora and LLMs tend to generate memorized text or new text that is similar to the input text. KGs are commonly built from authoritative and reliable data sources, enabling the generation of high-quality training data that align with human values, which is expected to enhance the security and reliability of LLMs.

\textbf{Computation-Intensive.} Training LLMs is computationally expensive, making it difficult to investigate their effectiveness with different techniques. The training process often requires thousands of GPUs and several weeks to complete. Moreover, LLMs are very computationally intensive and data hungry, making them difficult to deploy, especially in real-world applications where data and computing resources are limited. Through the integration of KGs, smaller LLMs have the potential to outperform larger ones, thereby reducing the cost associated with LLM deployment and application \cite{TOG}.

\textbf{Insufficient Interpretability.} Interpretability refers to how easily humans can comprehend a model's predictions, which is an essential gauge of the model's trustworthiness. LLMs are widely acknowledged as black boxes with opaque decision-making processes, making them challenging to interpret. KGs can be used to understand the knowledge learned by LLMs and interpret the reasoning process of LLMs, consequently enhancing the interpretability of LLMs \cite{swamy2021interpreting}.


Overall, LLMs have made noteworthy advancements and are considered a prototype of an artificial general intelligence system at its early stages. However, despite their ability to produce fluent and coherent text, they still encounter many obstacles. Among these obstacles, their struggle in recalling and accurately applying factual knowledge presents the primary challenge, and diminishes their ability to reason and accomplish knowledge-grounded tasks proficiently.



\section{KGPLMs}\label{sec:3}


In light of the limitations posed by poor factual knowledge modeling ability, researchers have proposed incorporating knowledge into PLMs to improve their performance. In recent years, various KGPLMs have been proposed, which can be categorized into before-training enhancement, during-training enhancement, and post-training enhancement methods according to the stage at which KGs participate in pre-training, as illustrated in Fig.~\ref{fig:kgplm}. 


\begin{table*}[!ht]
    \caption{Summary of KGPLMs}
    \centering
    \begin{tabular}{lll}
    \hline
        \multicolumn{2}{c}{Method} & KGPLM \\ \hline
        \multirow{4}{*}{Before-training enhancement} & Expand input structures & K-BERT \cite{K-BERT}, CoLAKE \cite{CoLAKE}, Zhang \textit{et al.} \cite{zhang2020cn} \\ 
	  \specialrule{0em}{1pt}{1pt}
        ~ & Enrich input information & LUKE \cite{LUKE}, E-BERT \cite{E-BERT}, KALM \cite{KALM}, OAG-BERT \cite{OAG-BERT}, DKPLM \cite{DKPLM} \\
	  \specialrule{0em}{1pt}{1pt} 
        ~ & Generate new data & AMS \cite{AMS}, KGPT \cite{KGPT}, KGLM \cite{KGLM}, ATOMIC \cite{ATOMIC}, KEPLER \cite{KEPLER} \\ 
	  \specialrule{0em}{1pt}{1pt}
        ~ & Optimize word masks & ERNIE \cite{ERNIE}, WKLM \cite{WKLM}, GLM \cite{GLM} \\ \hline
        \multirow{7}{*}{During-training enhancement} & Incorporate knowledge encoders & ERNIE \cite{ERNIE}, ERNIE 3.0 \cite{ERNIE-3}, BERT-MK \cite{BERT-MK}, CokeBERT \cite{CokeBERT}, JointLK \cite{JointLK}, \\ 
	   ~ & ~ & KET \cite{KET}, Liu et al. \cite{liu-etal-2022-relational}, QA-GNN \cite{QA-GNN}, GreaseLM \cite{GreaseLM}, KLMo \cite{KLMo} \\ 
	  \specialrule{0em}{1pt}{1pt}
        ~ & Insert knowledge encoding layers & KnowBERT \cite{KnowBERT}, K-BERT \cite{K-BERT}, CoLAKE \cite{CoLAKE}, JAKET \cite{JAKET}, KGBART \cite{KGBART} \\
	  \specialrule{0em}{1pt}{1pt}
        ~ & Add independent adapters & K-Adapter \cite{K-Adapter}, OM-ADAPT \cite{OM-ADAPT}, DAKI-ALBERT \cite{DAKI-ALBERT}, CKGA \cite{CKGA} \\ 
	  \specialrule{0em}{1pt}{1pt}
        ~ & Modify the pre-training task & ERNIE \cite{ERNIE}, LUKE \cite{LUKE}, OAG-BERT \cite{OAG-BERT}, WKLM \cite{WKLM}, SenseBERT \cite{SenseBERT}, \\
        ~ & ~ & ERICA \cite{ERICA}, SentiLARE \cite{SentiLARE}, GLM \cite{GLM}, KEPLER \cite{KEPLER}, JAKET \cite{JAKET}, \\
        ~ & ~ & ERNIE 2.0 \cite{ERNIE-2}, ERNIE 3.0 \cite{ERNIE-3}, DRAGON \cite{DRAGON}, LRLM \cite{LRLM} \\ \hline
        \multirow{3}{*}{Post-training enhancement} & Fine-tune PLMs with knowledge & KALA \cite{KALA}, KeBioSum \cite{KeBioSum}, KagNet \cite{KagNet}, BioKGLM \cite{BioKGLM}, \\
        ~ & ~ & Chang \textit{et al.} \cite{chang-etal-2020-incorporating} \\ 
	  \specialrule{0em}{1pt}{1pt}
        ~ & Generate knowledge-based prompts & Chang \textit{et al.} \cite{chang-etal-2020-incorporating}, Andrus \textit{et al.} \cite{andrus2022enhanced}, KP-PLM \cite{KP-PLM} \\ \hline
    \end{tabular}
    \label{tb:KGPLM-methods}
\end{table*}

\subsection{Before-training Enhancement KGPLMs}

\begin{figure}
    \centering
    \includegraphics[width=0.5\textwidth]{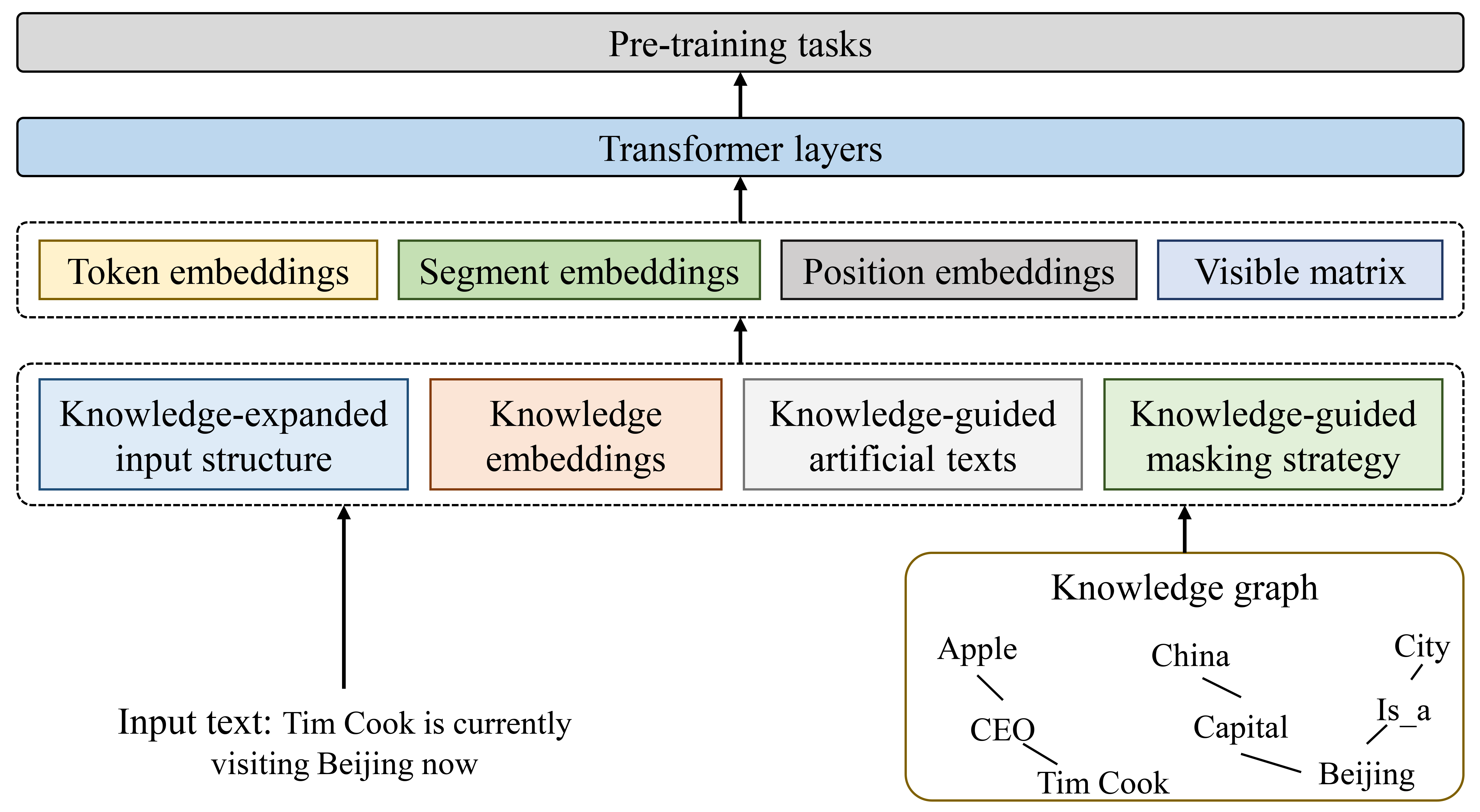}
    \caption{Main framework of before-training enhancement KGPLMs.}
    \label{fig:before-training}
\end{figure}

There are two challenges when integrating the knowledge from KGs into PLMs: heterogeneous embedding space and knowledge noise. The first challenge arises from the heterogeneity between text and KG. The second challenge occurs when unrelated knowledge diverts the sentence from its correct meaning. Before-training enhancement methods resolve these issues by unifying text and KG triples into the same input format, the framework of which is shown in Fig.~\ref{fig:before-training}. Existing studies propose diverse approaches to achieve this goal, including expanding input structures, enriching input information, generating new data, and optimizing word masks.

\textbf{Expand Input Structures.} Some methods expand the input text into graph structure to merge the structured knowledge of KGs and then convert the merged graph into text for PLM training. For example, K-BERT \cite{K-BERT} converts texts to sentence trees to inject related triples by fusing them with KG subgraphs and introduces soft-position and visible matrix to overcome the problem of knowledge noise. Moreover, it proposes mask-self-attention, an extension of self-attention, to prevent erroneous semantic alterations by taking advantage of the sentence structure information. Formally, the output from mask-self-attention is computed as:
\begin{equation}
    h=softmax(\frac{\mathbf{QK}^{T}+\mathbf{M}}{\sqrt{d_{k}}} )\mathbf{V},
\end{equation}
where $\mathbf{M}$ is the visible matrix. CoLAKE \cite{CoLAKE} addresses the heterogeneous embedding space challenge by combining knowledge context and language context into a unified word-knowledge graph. Zhang \textit{et al.} \cite{zhang2020cn} employed ConceptNet as the knowledge source and improved the visible matrix to control the information flow, which further improved the performance of K-BERT. 





\textbf{Enrich Input Information.} Instead of merging data from texts and KGs, some studies incorporate entities as auxiliary information by combining their embeddings with text embeddings. LUKE \cite{LUKE} introduces entity type embedding to indicate that the corresponding token in a sentence is an entity, and trains the model with the masked entity prediction task in addition to the MLM task. Further, it extends the Transformer encoder using an entity-aware self-attention mechanism to simultaneously handle both types of tokens.  E-BERT \cite{E-BERT} aligns entity embeddings with wordpiece vectors through an unconstrained linear mapping matrix and feeds the aligned representations into BERT as if they were wordpiece vectors. KALM \cite{KALM} signals the existence of entities to the input of the encoder in pre-training using an entity-extended tokenizer and adds an entity prediction task to train the model. Liu \textit{et al.} \cite{OAG-BERT} proposed OAG-BERT, a unified backbone language model for academic knowledge services, which integrates heterogeneous entity knowledge and scientific corpora in an open academic graph. They designed an entity type embedding to differentiate various entity types and used a span-aware entity masking strategy for MLM over entity names with different lengths. Besides, they designed the entity-aware 2D positional encoding to incorporate the entity span and sequence order information. Zhang \textit{et al.} \cite{DKPLM} decomposed the knowledge injection process of PLMs into pre-training, fine-tuning, and inference stages, and proposed DKPLM, which injects knowledge only during pre-training. Specifically, DKPLM detects long-tail entities according to their semantic importance in both texts and KGs and replaces the representations of detected long-tail entities with the representations of the corresponding knowledge triples generated by shared PLM encoders. The most-commonly used knowledge embedding model is TransE \cite{TransE}, which learns entity and relation representations by minimizing the following loss function:
\begin{equation}
    \mathcal{L} _{\mathrm{KE} } =-\left \| \mathbf{e} _{h}+\mathbf{r} -\mathbf{e} _{t} \right \|_{2}^{2},
\end{equation}
where $\mathbf{e} _{h}$ and $\mathbf{e} _{t}$ are the embeddings of the head and tail entities, while $\mathbf{r}$ is the representation of the relation.


\textbf{Generate New Data.} There are also some studies that inject knowledge into PLMs by generating artificial text based on KGs. For example, AMS \cite{AMS} constructs a commonsense-related question answering dataset for training PLMs based on an align-mask-select method. Specifically, it aligns sentences with commonsense knowledge triples, masks the aligned entities in the sentences and treats the masked sentences as questions. In the end, it selects several entities from KGs as distractor choices and trains the model to determine the correct answer. KGPT \cite{KGPT} crawls sentences with hyperlinks from Wikipedia and aligns the hyperlinked entities to the KG Wikidata to construct the knowledge-grounded corpus KGText. KGLM \cite{KGLM} constructs the Linked WikiText-2 dataset by aligning texts in WikiText-2 and entities in Wikidata. ATOMIC \cite{ATOMIC} organizes the inference knowledge in 877K textual descriptions into a KG and trains a PLM with a conditional sequence generation problem that encourages the model to generate the target sequence given an event phrase and an inference dimension. KEPLER \cite{KEPLER} constructs a large-scale KG dataset with aligned entity descriptions from its corresponding Wikipedia pages for training KGPLMs.


\textbf{Optimize Word Masks.} MLM is the most commonly used pre-training task in PLMs, and the number and distribution of masks have a substantial influence on the performance of PLMs \cite{wettig2022should}. However, the random masking method may break the correlation between consecutive words, making it difficult for PLMs to learn semantic information. To address this issue, a few studies have proposed replacing the random masking strategy with a knowledge masking strategy that selects mask targets based on the knowledge from KGs, forcing models to learn enough knowledge to accurately predict the masked contents. For instance, ERNIE \cite{ERNIE} recognizes named entities in texts and aligns them with their corresponding entities in KGs. It then randomly masks entities in the input text and trains the model to select their counterparts in KGs. In WKLM \cite{WKLM}, entity mentions in the original texts are substituted with entities of identical types, and the model is trained to differentiate accurate entity mentions from those that are corrupted, which effectively improves its fact completion performance. GLM \cite{GLM} reformulates the MLM objective to an entity-level masking strategy that identifies entities and selects informative ones by considering both document frequency and mutual reachability of the entities detected in the text.

Before-training enhancement methods can improve the semantic standardization and structural level of the corpus, which is helpful for improving the reasoning ability of PLMs \cite{bi2023program} without improving the model size and training time. Besides, the training data enhanced by KGs can better describe commonsense knowledge, which helps to improve LLMs' commonsense knowledge modeling ability. These methods are more suitable for those domains without sufficient training corpus and can effectively improve LLMs' performance and generalization ability in such domains. However, before-training enhancement processing requires additional computational resources and time, making the pre-training process more complex and cumbersome. Besides, it may introduce noise, which can have a negative impact on LLMs' training.


\subsection{During-training Enhancement KGPLMs}

\begin{figure*}
    \centering
    \includegraphics[width=0.9\textwidth]{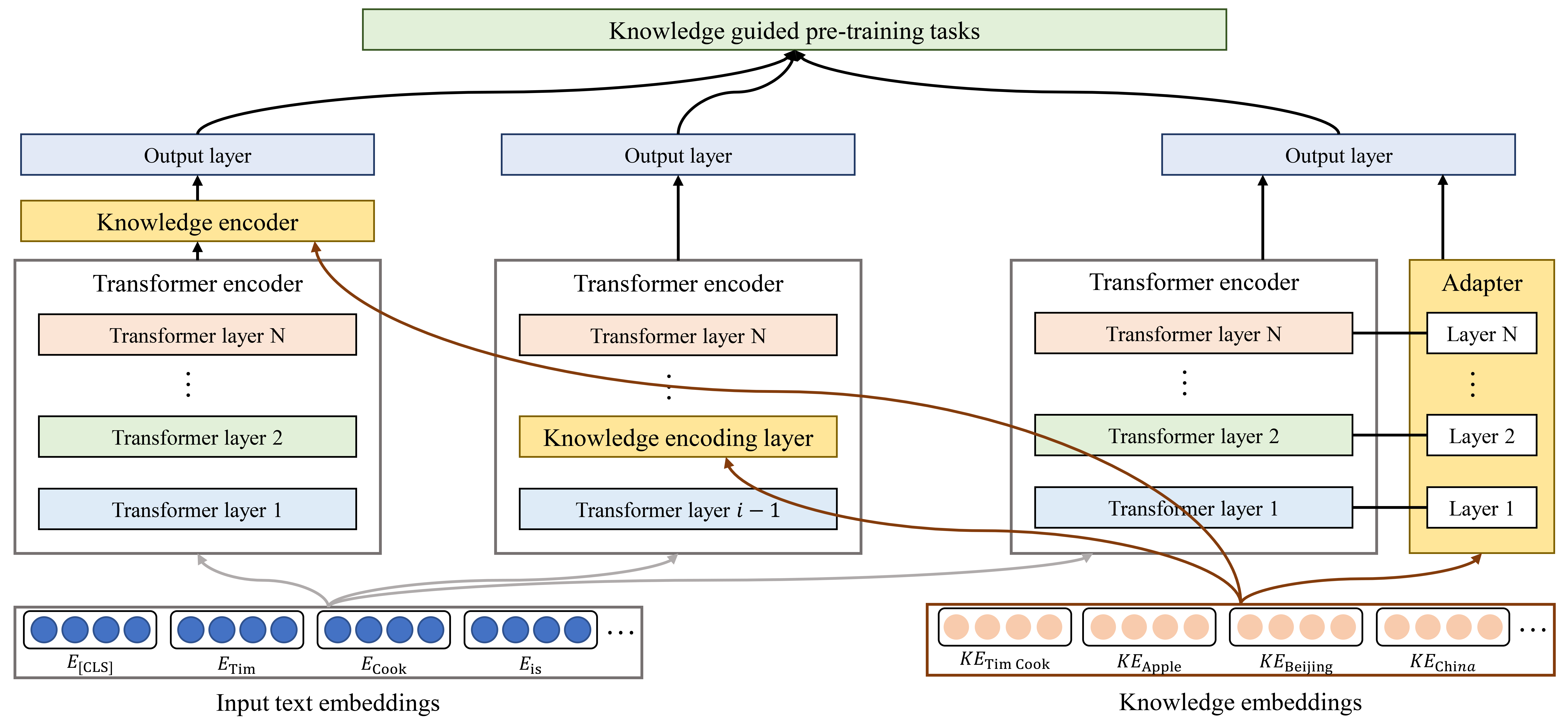}
    \caption{Main framework of during-training enhancement KGPLMs.}
    \label{fig:in-training}
\end{figure*}

During-training enhancement methods enable PLMs to learn knowledge directly during training by improving their encoder and training task. Since plain PLMs cannot process text sequences and structured KG simultaneously, some studies have proposed incorporating knowledge encoders or external knowledge modules to enable learning from both text and KGs concurrently. Existing during-training enhancement KGPLMs can be divided into incorporating knowledge encoders, inserting knowledge encoding layers, adding independent adapters, and modifying the pre-training task, as shown in Fig.~\ref{fig:in-training}.



\textbf{Incorporate Knowledge Encoders.} ERNIE \cite{ERNIE} integrates a knowledge encoder to incorporate KG information, which takes two types of input: the token embedding and the concatenation of the token and entity embeddings. Building on ERNIE, ERNIE 3.0 \cite{ERNIE-3} builds a few task-specific modules upon the universal representation module to enable easy customization of the model for natural language understanding and generation tasks. BERT-MK \cite{BERT-MK} utilizes a graph contextualized knowledge embedding module to learn knowledge in subgraphs and incorporates the learned knowledge into the language model for knowledge generalization. CokeBERT \cite{CokeBERT} utilizes three modules to select contextual knowledge and embed knowledge context, where the text encoder computes embeddings for the input text, the knowledge context encoder dynamically selects knowledge context based on textual context and computes knowledge embeddings, while the knowledge fusion encoder fuses textual context and knowledge context embeddings for better language understanding. JointLK \cite{JointLK} performs joint reasoning between PLM and a graph neural network (GNN) through a dense bidirectional attention module to effectively fuse and reason over question and KG representations. KET \cite{KET} interprets contextual utterances using hierarchical self-attention and dynamically leverages external commonsense knowledge using a context-aware affective graph attention mechanism to detect emotions in textual conversations. Liu \textit{et al.} \cite{liu-etal-2022-relational} proposed a memory-augmented approach to condition a PLM on a KG, which represents the KG as a set of relation triples and retrieves pertinent relations for a given context to enhance text generation. QA-GNN \cite{QA-GNN} uses a PLM to estimate the importance of nodes to identify relevant knowledge from large KGs, and combines the QA context and KG to form a joint graph. Then, it mutually updates the representations of QA context and KG through graph-based message passing to perform joint reasoning. GreaseLM \cite{GreaseLM} integrates embeddings from a PLM and a GNN through several layers of modality interaction operations. KLMo \cite{KLMo} explicitly models the interaction between entity spans in texts and all entities and relations in a contextual KG using a novel knowledge aggregator. 


\textbf{Insert Knowledge Encoding Layers.} Some methods insert additional knowledge encoding layers in the middle of PLMs or adjust the encoding mechanism to enable PLMs to process knowledge. For instance, KnowBERT \cite{KnowBERT} incorporates a knowledge attention recontextualization module to integrate multiple KGs into a PLM. It explicitly models entity spans within the input text and uses an entity linker to retrieve relevant entity embeddings from the KG. These retrieved embeddings are then utilized to create knowledge-enhanced entity-span embeddings. K-BERT \cite{K-BERT} changes the Transformer encoder to a mask-Transformer, which takes soft-position and visible matrix as input to control the influence of knowledge and avoid the knowledge noise issue. CoLAKE \cite{CoLAKE} slightly modifies the embedding layer and encoder layers of Transformer to adapt to input in the form of word-knowledge graph. This graph combines the knowledge context and language context into a unified data structure. JAKET \cite{JAKET} decomposes the encoder of a PLM into two modules, with the first providing embeddings for both the second and KG, while the second module takes text and entity embeddings to produce the final representation. KGBART \cite{KGBART} follows the BART architecture but replaces the traditional Transformer with an effective knowledge graph-augmented Transformer to capture relations between concept sets, where KGs serve as additional inputs to the graph attention mechanism.


\textbf{Add Independent Adapters.} Some methods add independent adapters to process knowledge, which are easy to train and whose training process does not affect the parameters of the original PLM. For instance, K-Adapter \cite{K-Adapter} enables the injection of various types of knowledge by training adapters independently on different tasks. This approach facilitates the continual fusion of knowledge. OM-ADAPT \cite{OM-ADAPT} complements BERT's distributional knowledge by incorporating conceptual knowledge from ConceptNet and the corresponding Open Mind Common Sense corpus through adapter training. This approach avoids the expensive computational overhead of joint pre-training, as well as the problem of catastrophic forgetting associated with post-hoc fine-tuning. DAKI-ALBERT \cite{DAKI-ALBERT} proposes pre-training knowledge adapters for specific domain knowledge sources and integrating them through an attention-based knowledge controller to enhance PLMs with enriched knowledge. CKGA \cite{CKGA} introduces a novel commonsense KG-based adapter for sentiment classification tasks, which utilizes a PLM to encode commonsense knowledge and extracts corresponding knowledge with a GNN.

\textbf{Modify the Pre-training Task.} Several studies attempt to incorporate knowledge into PLMs by modifying the pre-training tasks. The most commonly used method is to change MLM to masked entity modeling (MEM) based on entities marked in texts. Examples of such methods include ERNIE \cite{ERNIE}, LUKE \cite{LUKE}, OAG-BERT \cite{OAG-BERT}, WKLM \cite{WKLM}, etc. SenseBERT \cite{SenseBERT} directly applies weak supervision at the word sense level, which trains a PLM to predict not only masked words but also their WordNet supersenses. ERICA \cite{ERICA} defines two novel pre-training tasks to explicitly model relational facts in texts through contrastive learning, in which the entity discrimination task trains the model to distinguish tail entities while the relation discrimination task is designed to train the model to distinguish the proximity between two relations. SentiLARE \cite{SentiLARE} introduces a context-aware sentiment attention mechanism to determine the sentiment polarity of each word based on its part-of-speech tag by querying SentiWordNet. It also proposes a novel pre-training task called label-aware masked language model to build knowledge-aware language representations. GLM \cite{GLM} introduces a KG-guided masking scheme and then employs KGs to obtain distractors for masked entities and uses a novel distractor-suppressed ranking objective to optimize the model.


Other methods utilize the multi-task learning mechanism to integrate knowledge representation learning with the training of PLMs, simultaneously optimizing knowledge representation and model parameters. KEPLER \cite{KEPLER} employs a shared encoder to encode texts and entities into a unified semantic space, while simultaneously optimizing knowledge embedding and MLM objectives. JAKET \cite{JAKET} jointly models KG and language using two modules, in which the language module and knowledge module mutually assist each other through embeddings. Building upon ERNIE, ERNIE 2.0 \cite{ERNIE-2} proposes a continual multi-task learning framework that extracts valuable lexical, syntactic, and semantic information. ERNIE 3.0 \cite{ERNIE-3} combines auto-regressive and auto-encoding networks to process multiple pre-training tasks at both language and knowledge levels. DRAGON \cite{DRAGON} uses a cross-modal encoder that bidirectionally exchanges information between text tokens and KG nodes to produce fused representations and trains this encoder by unifying two self-supervised reasoning tasks: MLM and KG link prediction. LRLM \cite{LRLM} parameterizes the joint distribution over the words in a text and the entities therein, leveraging KGs through relations when modeling text.

During-training enhancement methods can adaptively incorporate external knowledge while learning parameters, often leading to improved performance on various downstream tasks. Moreover, they allow for customization to specific domains or tasks by introducing special information or modules. However, they may increase training time as they typically improve the parameter size and could be limited by the scope of knowledge included in the training data. Moreover, with more complex architecture and more parameters, LLMs are more susceptible to overfitting and require more training to maintain generalization. During-training enhancement methods are more suitable for those scenarios that require dealing with multiple complex tasks, and they often perform better on knowledge-grounded tasks than other methods.

\subsection{Post-training Enhancement KGPLMs}

\begin{figure}
    \centering
    \includegraphics[width=0.5\textwidth]{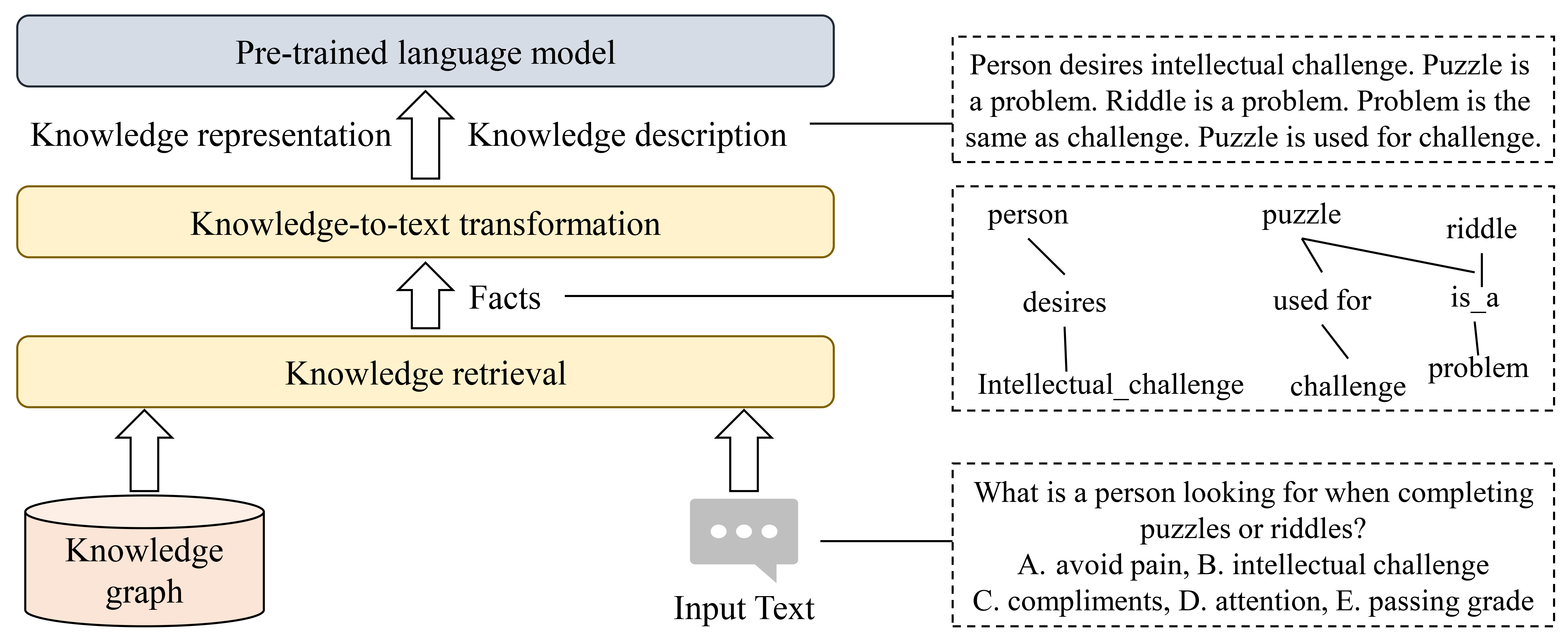}
    \caption{Main framework of post-training enhancement KGPLMs.}
    \label{fig:post-training}
\end{figure}

Post-training enhancement methods typically inject domain-specific knowledge into PLMs through fine-tuning them on additional data and tasks, which improves the model's performance on specific domain tasks. Additionally, with the rapid development of prompt learning \cite{prompt}, several recent investigations have proposed automatically generating prompts to improve the outputs of PLMs. The main framework of post-training enhancement KGPLMs is shown in Fig.~\ref{fig:post-training}.

\textbf{Fine-tune PLMs with Knowledge.} KALA \cite{KALA} modulates PLMs' intermediate hidden representations with domain knowledge, which largely outperforms adaptive pre-training models while still being computationally efficient. KeBioSum \cite{KeBioSum} investigates the integration of generative and discriminative training techniques to fuse knowledge into knowledge adapters. It applies adapter fusion to effectively incorporate these knowledge adapters into PLMs for the purpose of fine-tuning biomedical text summarization tasks. KagNet \cite{KagNet} proposes a textual inference framework for answering commonsense questions, which effectively utilizes KGs to provide human-readable results via intermediate attention scores. BioKGLM \cite{BioKGLM} presents a post-training procedure between pre-training and fine-tuning and uses diverse knowledge fusion strategies to facilitate the injection of KGs. Chang \textit{et al.} \cite{chang-etal-2020-incorporating} proposed attentively incorporating retrieved tuples from KGs to incorporate commonsense knowledge during fine-tuning.


\textbf{Generate Knowledge-based Prompts.} Bian \textit{et al.} \cite{bian2021benchmarking} presented a knowledge-to-text framework for knowledge-enhanced commonsense question-answering. It transforms structured knowledge into textual descriptions and utilizes machine reading comprehension models to predict answers by exploiting both original questions and textural knowledge descriptions. Andrus \textit{et al.} \cite{andrus2022enhanced} proposed using open information extraction models with rule-based post-processing to construct a custom dynamic KG. They further suggested utilizing few-shot learning with GPT-3 to verbalize extracted facts from the KG as natural language and incorporate them into prompts. KP-PLM \cite{KP-PLM} constructs a knowledge sub-graph from KGs for each context and adopts multiple continuous prompt rules to transform the knowledge sub-graph into natural language prompts. Furthermore, it leverages two novel knowledge-aware self-supervised tasks: prompt relevance inspection and masked prompt modeling, to optimize the model.

\begin{table}[!ht]
\caption{Performance improvement of some KGPLMs on different evaluation tasks compared with BERT}
\setlength\tabcolsep{3pt}
    \centering
    \begin{tabular}{cccc}
    \hline
        KGPLM & Entity typing & Relation classification & Question answering \\ \hline
        CoLAKE & 2.8 & 5.6 & — \\ 
        LUKE & 4.6 & 6.7 & 19.2 \\
        KEPLER & 2.6 & 6 & — \\ 
        ERNIE & 2 & 3.4 & — \\ 
        CokeBERT & 1.3 & 2.7 & — \\ 
        K-Adapter & 4.1 & 1.9 & 5.4 \\ 
        ERICA & 4.4 & 2.2 & 1.5 \\ 
        KP-PLM & 4.6 & 3.8 & — \\ \hline
    \end{tabular}
\label{tb:KGPLM-performance}
\end{table}

Post-training enhancement methods are low-cost and easy to implement, which can effectively improve LLMs' performance on specific tasks. Besides, these methods can guide LLMs to generate text of specific styles and improve the quality and security of LLMs' output. Therefore, post-training enhancement methods are more suitable for domain-specific tasks and text generation scenarios that require sensitive information filtering and risk control. However, the labeling of fine-tuning data and the design of prompts rely on prior knowledge and external resources. If there is a lack of relevant prior knowledge, the optimization effect may be limited. Moreover, these methods may impose certain limitations on the flexibility of LLMs' generations. The generated text may be constrained by prompts and may not be able to be fully freely created.

\subsection{Effectiveness and Efficiency of KGPLMs}

Most KGPLMs are designed for knowledge-grounded tasks. To evaluate their effectiveness in knowledge modeling, we report their performance on three knowledge-grounded tasks: entity typing, relation classification, and question answering. Table~\ref{tb:KGPLM-performance} provides a summary of KGPLMs and their respective improvements over the unenhanced BERT. The reported metric is F1-score. In Table~\ref{tb:KGPLM-performance}, the performances of these models on all tasks are higher than BERT, indicating that KGs enhance their knowledge modeling ability.

\begin{table}[!ht]
\caption{The running time of BERT and different KGPLMs}
\setlength\tabcolsep{12pt}
    \centering
    \begin{tabular}{cccc}
    \hline
        Model & Pre-training & Fine-tuning & Inference \\ \hline
        BERT & 8.46 & 6.76 & 0.97 \\
        RoBERTa & 9.60 & 7.09 & 1.55 \\ 
        ERNIE & 14.71 & 8.19 & 1.95 \\ 
        KEPLER & 18.12 & 7.53 & 1.86 \\
        CoLAKE & 12.46 & 8.02 & 1.91 \\ 
        DKPLM & 10.02 & 7.16 & 1.61 \\ \hline
    \end{tabular}
\label{tb:KGPLM-time}
\end{table}

Typically, the incorporation of knowledge from KGs would lead to a larger parameter size compared with the base PLM. Consequently, the pre-training, fine-tuning and inference time of plain PLMs are consistently shorter than KGPLMs. As the statistical data shown in Table~\ref{tb:KGPLM-time}, due to these KGPLMs injecting the knowledge encoder module into PLMs, their running time of the three stages are consistently longer than BERT. However, with the incorporation of external knowledge, KGPLMs are easier to be trained with higher performance. For example, KALM with 775M parameters even performs better than GPT-2 on some downstream tasks \cite{KALM}, whose parameter size is 1.5B. This implies that we can obtain a satisfactory model with smaller parameter size and fewer training resources.

\section{Applications of KGPLMs}

\begin{figure*}
    \centering
    \includegraphics[width=0.9\textwidth]{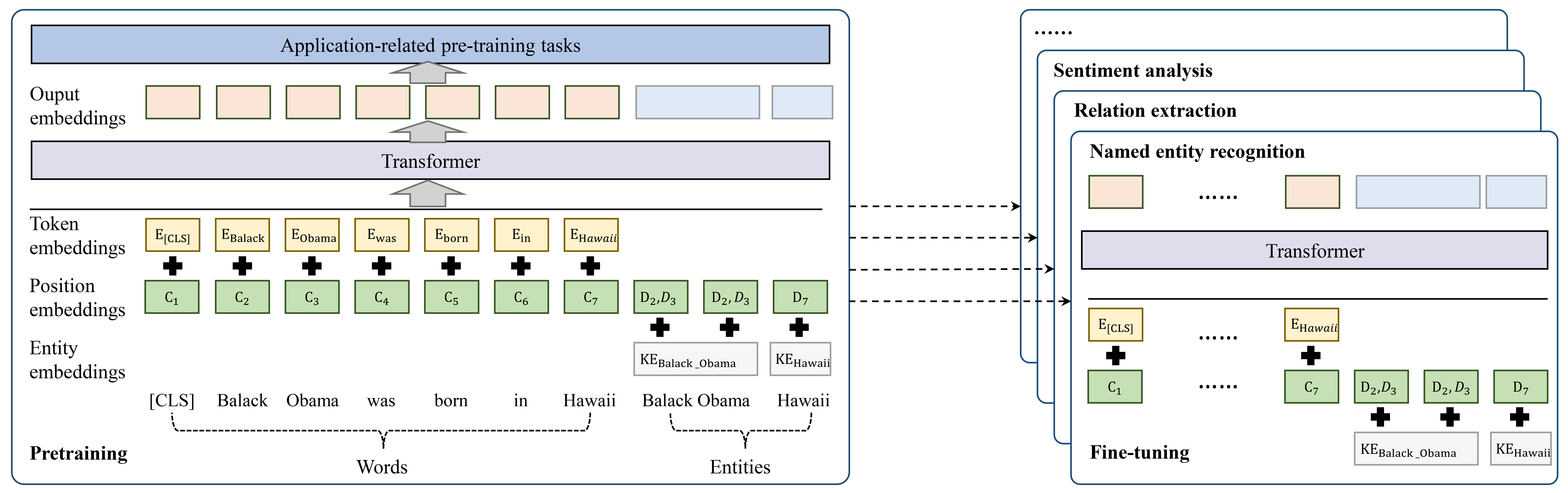}
    \caption{The framework for KGPLMs to realize various applications.}
    \label{fig:application}
\end{figure*}

KGPLMs outperform traditional PLMs in capturing factual and relational information, exhibiting stronger language understanding and generation abilities. These advantages lead to improved performance across a range of downstream applications. By employing diverse pre-training tasks and fine-tuning PLMs for specific applications, as illustrated in Fig.~\ref{fig:application}, KGPLMs have been successfully leveraged for multiple tasks.


\textbf{Named Entity Recognition.} Named entity recognition (NER) aims to identify entities with specific meanings from text, such as names of persons, places, and organizations. PLMs have successfully improved state-of-the-art word representations and demonstrated effectiveness on the NER task by modeling context information \cite{giorgi2019end}. However, these models are trained to predict correlations between tokens, ignoring the underlying meanings behind them and the complete semantics of entities that consist of multiple tokens \cite{KeBioLM}. Previous work has already regarded NER as a knowledge intensive task and improved PLMs' NER performance by incorporating external knowledge into PLMs \cite{seyler-etal-2018-study}. Therefore, researchers have developed KGPLMs for NER, which can leverage additional information beyond the training corpus for better performance, especially in domain-specific tasks where the training samples are often insufficient. For example, He \textit{et al.} \cite{KAWR} incorporated prior knowledge of entities from an external knowledge base into word representations and introduced a KG augmented word representation framework for NER. Some other KGPLMs like K-BERT \cite{K-BERT} and ERNIE \cite{ERNIE} also demonstrate their superiority on diverse NER datasets.

\textbf{Relation Extraction.} Relation extraction involves distinguishing semantic relationships between entities and classifying them into predefined relation types. Although PLMs have improved the efficacy of relation extraction to some extent, when applied to small-scale and domain-specific texts, there is still a lack of information learning \cite{3578781}. To address this limitation, several studies have suggested injecting prior knowledge from KGs into PLMs. KGPLMs have been demonstrated to be more effective than plain PLMs in relation extraction \cite{SPOT}. For example, Roy \textit{et al.} \cite{roy-pan-2021-incorporating} proposed merging KG embeddings with BERT to improve its performance on clinical relation extraction. BERT-MK \cite{BERT-MK} also demonstrates the effectiveness of KGPLMs on biomedical relation extraction. In addition to the biomedical field, KGPLMs such as KEPLER \cite{KEPLER} and JAKET \cite{JAKET} are also commonly applied to public domain relation extraction tasks.


\textbf{Sentiment Analysis.} Sentiment analysis aims to analyze whether the emotions expressed in the text are positive, negative, or neutral. Recently, sentiment analysis has made remarkable advances with the help of PLMs, which achieve state-of-the-art performance on diverse benchmarks. However, current PLMs focus on acquiring semantic information through self-supervision techniques, disregarding sentiment-related knowledge throughout pre-training \cite{Sentix}. By integrating different types of sentiment knowledge into the pre-training process, the learned semantic representation would be more appropriate. For this reason, several KGPLMs have been applied to sentiment analysis, including SentiLARE \cite{SentiLARE}, KCF-PLM \cite{KCF-PLM}, and KET \cite{KET}, which have proven the effectiveness of injecting KGs into PLMs for sentiment analysis.


\textbf{Knowledge Graph Completion.} Due to the limitations in data quality and automatic extraction technology, KGs are often incomplete, and some relations between entities are missing \cite{9531531}. Therefore, the knowledge graph completion task, aiming at inferring missing relations and improving the completeness of KGs, has been widely investigated. Given the triumph of PLMs, some PLM-based methods are proposed for the knowledge graph completion task. Nonetheless, most of these methods concentrate on modeling the textual representation of factual triples while neglecting the underlying topological contexts and logical rules that are essential for KG modeling \cite{FTL-LM,9525301}. To address this challenge, some studies have suggested combining topology contexts and logical rules in KGs with textual semantics in PLMs to complete the KG. By integrating the structure information from KGs and the contextual information from texts, KGPLMs outperform those PLMs specifically designed for the KG completion task \cite{GLM}. We can also extract the knowledge-enhanced embeddings to predict the rationality of given triples \cite{KEPLER}.


\textbf{Question Answering.} Question answering systems need to choose the correct answers for the given questions, which must be able to access relevant knowledge and reason over it. Although PLMs have made remarkable achievements on many question answering tasks \cite{8525315}, they do not empirically perform well on structured reasoning. On the other hand, KGs are more suitable for structured reasoning and enable explainable predictions. Therefore, a few studies have proposed integrating PLMs with KGs to conduct structured reasoning and enable explainable predictions. Some methods incorporate KGs into PLMs while training them, such as QA-GNN \cite{QA-GNN} and WKLM \cite{WKLM}. Another line of research uses KGs to augment PLMs during answer inference. OreoLM \cite{OreoLM}, for example, incorporates a novel knowledge interaction layer into PLMs that interact with a differentiable knowledge graph reasoning module for collaborative reasoning. Here, PLMs guide KGs in walking towards desired answers while retrieved knowledge enhances PLMs. Experiments on common benchmarks illustrate that KGPLMs outperform traditional PLMs after KG incorporation.


\textbf{Natural Language Generation.} Natural language generation (NLG) serves as a fundamental building block for various applications in NLP, such as dialogue systems, neural machine translation, and story generation, and has been subject to numerous studies. Deep neural language models pre-trained on large corpus have caused remarkable improvements in multiple NLG benchmarks. However, even though they can memorize enough language patterns during pre-training, they merely capture average semantics of the data and most of them are not explicitly aware of domain-specific knowledge. Thus, when specific knowledge is required, contents generated by PLMs could be inappropriate. KGs, which store entity attributes and their relations, contain rich semantic contextual information. As a result, several studies have proposed incorporating KGs into PLMs to improve their NLG performance. For instance, Guan \textit{et al.} \cite{guan2020knowledge} proposed improving GPT-2 with structured knowledge by post-training the model using knowledge examples sourced from KGs. They aimed to supply additional crucial information for story generation. Ji \textit{et al.} \cite{ji2020language} proposed GRF, a generation model that performs multi-hop reasoning on external KGs, enriching language generation with KG-derived data. Experimental results indicate that KGPLMs outperform PLMs in story ending generation \cite{DICE}, abductive reasoning \cite{ege-RoBERTa}, and question answering \cite{QA-GNN}. 



\textbf{Industrial Applications.} KGPLMs have been applied in many real-world applications. Typical applications include chatbots, such as ERNIE Bot\footnote{https://yiyan.baidu.com/} from Baidu, Qianwen\footnote{https://qianwen.aliyun.com/} from Alibaba, and Bard\footnote{https://bard.google.com/} from Google, which incorporate KGs into PLMs to improve knowledge awareness while communicating with humans. Such applications have shown that KGPLMs can provide excellent language understanding and knowledge modeling abilities. PLMs have also been successfully applied in programming assistants, which can easily generate codes according to context or natural language prompts. However, there are still some issues encountered by PLM-based programming assistants, such as incorrect code recommendations and excessive reliance on code libraries. To tackle these challenges, GitHub and OpenAI released Copilot X\footnote{https://github.com/features/preview/copilot-x}, which incorporates KGs into the programming assistant to analyze the logical dependencies of the code and generate appropriate code recommendations. Aside from the above applications, KGPLMs are widely used in a variety of virtual assistants and search engines. Representatives of these applications include Xiaodu\footnote{https://dueros.baidu.com/en/index.html} from Baidu and PanGu\footnote{https://www.huaweicloud.com/product/pangu.html} from Huawei, which can respond to a broad range of queries like weather forecasts, singing songs, and navigation.


\section{Can LLMs Replace KGs?}

\begin{figure}
    \centering
    \includegraphics[width=0.5\textwidth]{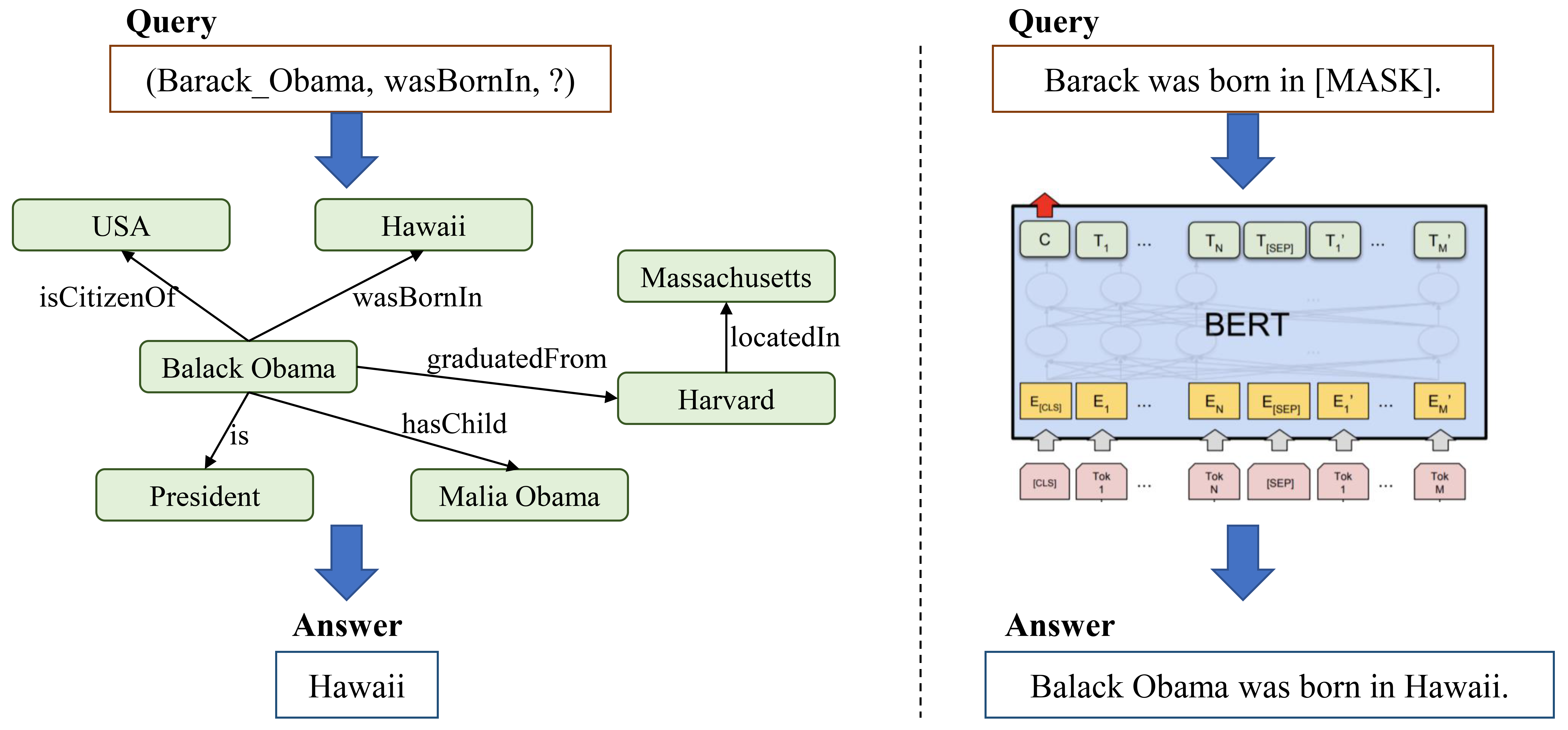}
    \caption{Querying KGs and PLMs for factual knowledge, in which the left part represents directly querying factual knowledge from KGs while the right part represents querying factual knowledge from PLMs by asking them to fill in masked tokens in sequences.}
    \label{fig:lmkb}
\end{figure}

Recent advancements in training PLMs on a large corpus have led to a surge of improvements for downstream NLP tasks. While primarily learning linguistic knowledge, PLMs may also store some relational knowledge present in the training data that enables them to answer complex queries. Although their knowledge cannot be directly queried like KGs, we can attempt to query them for factual knowledge by asking them to fill in masked tokens in sequences, as illustrated in Fig.~\ref{fig:lmkb}. Consequently, some researchers believe that parametric PLMs can replace symbolic KGs as knowledge bases \cite{alkhamissi2022review}. For example, Petroni \textit{et al.} \cite{LAMA} proposed LAMA, a knowledge probe consisting of cloze-style queries, to measure relational knowledge contained in PLMs. Their results show that PLMs contain relational knowledge and can recall stored facts without fine-tuning. Talmor \textit{et al.} \cite{talmor2020olmpics} developed eight cloze-style reasoning tasks to test the knowledge captured in BERT and RoBERTa. They found that different PLMs exhibit qualitatively different reasoning abilities and do not reason in an abstract manner but instead rely on context. Heinzerling and Inui \cite{heinzerling-inui-2021-language} evaluated PLMs' ability to store millions of entity facts and query these facts via experimental tests with three entity representations. Their experimental results provide a proof-of-concept for PLMs as knowledge bases.


Nevertheless, after conducting extensive experimental analyses of PLMs, some studies have reported that PLMs struggle to accurately recall relational facts, raising doubts about their viability as knowledge bases. A surge of benchmark datasets and tasks have been proposed to examine the knowledge embodied within PLMs. For example, Wang \textit{et al.} \cite{wang-etal-2019-make} released a benchmark to directly test a system's ability to differentiate natural language statements that make sense from those that do not. By comparing their performance with humans, they revealed that sense-making remains a technical challenge for PLMs. Sung \textit{et al.} \cite{sung2021can} created the BioLAMA benchmark that is comprised of 49K biomedical factual knowledge triples for probing biomedical PLMs. Their detailed analysis reveals that most PLMs' predictions are highly correlated with prompt templates without any subjects, hence producing similar results on each relation and hindering their capabilities to be used as biomedical knowledge bases. Wang \textit{et al.} \cite{wang2021can} constructed a new dataset of closed-book question answering and tested the BART's \cite{BART} ability to answer these questions. Experimental results show that it is challenging for BART to answer closed-book questions since it cannot remember training facts in high precision. Zhao \textit{et al.} \cite{LAMA-TK} introduced LAMA-TK, a dataset aimed at probing temporally-scoped knowledge. They investigated the capacity of PLMs for storing temporal knowledge that contains conflicting information and the ability to use stored knowledge for temporally-scoped knowledge queries. Their experimental results show that conflicting information poses great challenges to PLMs, which drops their storage accuracy and hinders their memorization of multiple answers. Kassner \textit{et al.} \cite{kassner2021multilingual} translated two established benchmarks into 53 languages to investigate the knowledge contained in the multilingual PLM mBERT \cite{mBERT}. They found that mBERT yielded varying performance across languages. The above studies have proven that PLMs still face challenges in accurately storing knowledge, dealing with knowledge diversity, and retrieving correct knowledge to solve corresponding tasks. Additionally, Cao \textit{et al.} \cite{cao-etal-2021-knowledgeable} conducted a comprehensive investigation into the predictive mechanisms of PLMs across various extraction paradigms. They found that previous decent performance of PLMs mainly owes to the biased prompts which overfit dataset artifacts. AlKhamissi \textit{et al.} \cite{alkhamissi2022review} suggested five essential criteria that PLMs should meet in order to be considered proficient knowledge bases: access, edit, consistency, reasoning, and explainability and interpretability, and found that PLMs do not perform as well as KGs in terms of consistency, reasoning, and interpretability. They also reviewed the literature with respect to the five aspects and revealed that the community still has a long way to go to enable PLMs to serve as knowledge bases despite some recent breakthroughs. These studies raise doubts about PLMs' potential as knowledge bases and underscore the need for further research in this area.

Despite the fact that larger-sized LLMs seem to possess more fundamental knowledge of the world, their learned encyclopedic facts and common sense properties of objects are still unreliable. Furthermore, they have limited capabilities in inferring relationships between actions and events \cite{chang2023language}. The ability of LLMs to predict facts is also significantly dependent on specific prompt templates and the included entities \cite{cao2022can}. This owes to the fact that LLMs mainly rely on simple heuristics with most predictions correlated to co-occurrence frequencies of the target word and words in the prompt. Additionally, the accuracy of their predictions is highly reliant on the frequency of facts in the pre-training corpus \cite{kandpal2022large}.

To summarize, LLMs and KGs have their respective advantages and disadvantages. KGs lack the flexibility that LLMs offer, as KGs require substantial human effort to build and maintain, while LLMs provide more flexibility through unsupervised training on a large corpus. However, KGs are easier to access and edit, and have better consistency, reasoning ability, and interpretability. First, factual knowledge in KGs is often easily accessed through manual query instructions. In contrast, LLMs cannot be queried explicitly, as the knowledge is implicitly encoded in their parameters. Second, the triplets in KGs can be directly added, modified, and deleted. However, editing a specific fact in LLMs is not straightforward, since facts in LLMs cannot be directly accessed. To enable LLMs to learn up-to-date, correct, and unbiased knowledge, the whole model needs to be retrained on updated data, which is expensive and inflexible. Third, KGs are built with consistency in mind, and various algorithms have been proposed to eliminate conflicts that arise in KGs. On the other hand, LLMs may be inconsistent, as they may yield different answers to the same underlying factual questions. Fourth, it can be simple to follow the path of reasoning in KGs, while LLMs perform poorly on relational reasoning tasks. Finally, KGs have a clear reasoning path, so their outputs are easy to interpret. However, as typical black-box models, knowledge is hard to be identified by simply looking at LLMs' outputs.


Although current LLMs face limitations in directly serving as knowledge bases, they contribute to constructing KGs that explicitly express their stored knowledge. One approach is to utilize LLMs as an information extraction tool to improve the accuracy of NER and relation extraction. Another way is to extract symbolic KGs from LLMs using prompts. For example, Hao \textit{et al.} \cite{BertNet} proposed a novel framework to automatically construct KGs from LLMs that generates diverse prompts, searches for consistent outputs, and performs efficient knowledge search. Bosselut \textit{et al.} \cite{COMET} proposed a fine-tuned generative LLM for the automatic construction of commonsense KGs that generates tail entities based on given head entities and relations. These approaches demonstrate the potential of leveraging LLMs for effective KG construction.



To conclude, LLMs still face challenges in remembering large amounts of complex knowledge and retrieving the required information accurately. There are multiple aspects in which LLMs need to excel to qualify as comprehensive knowledge bases. On the other hand, KGs and LLMs complement each other, enhancing overall performance. Therefore, enhancing LLMs with KGs can significantly improve their performance on knowledge-grounded tasks.

\section{Enhancing LLMs with KGs}

In the preceding sections, we have analyzed and compared existing KGPLMs. Despite demonstrating proficiency in a wide range of NLP tasks, the complexity of knowledge and language continues to pose unresolved challenges for KGPLMs. Furthermore, despite substantial improvements in generated text quality and learned facts with models scaling beyond 100B parameters, LLMs are still prone to unfactual responses and commonsense errors. Their predictions are highly dependent on input text, and minor variations in phrasing and word choice can lead to such errors. One potential solution is to enhance LLMs with KGs to improve their learning of factual knowledge, a topic that has not been thoroughly studied yet. Thus, we propose to enhance LLMs with KGs using techniques utilized by KGPLMs to achieve fact-aware language modeling.


\subsection{Overall Framework}

\begin{figure}
    \centering
    \includegraphics[width=0.5\textwidth]{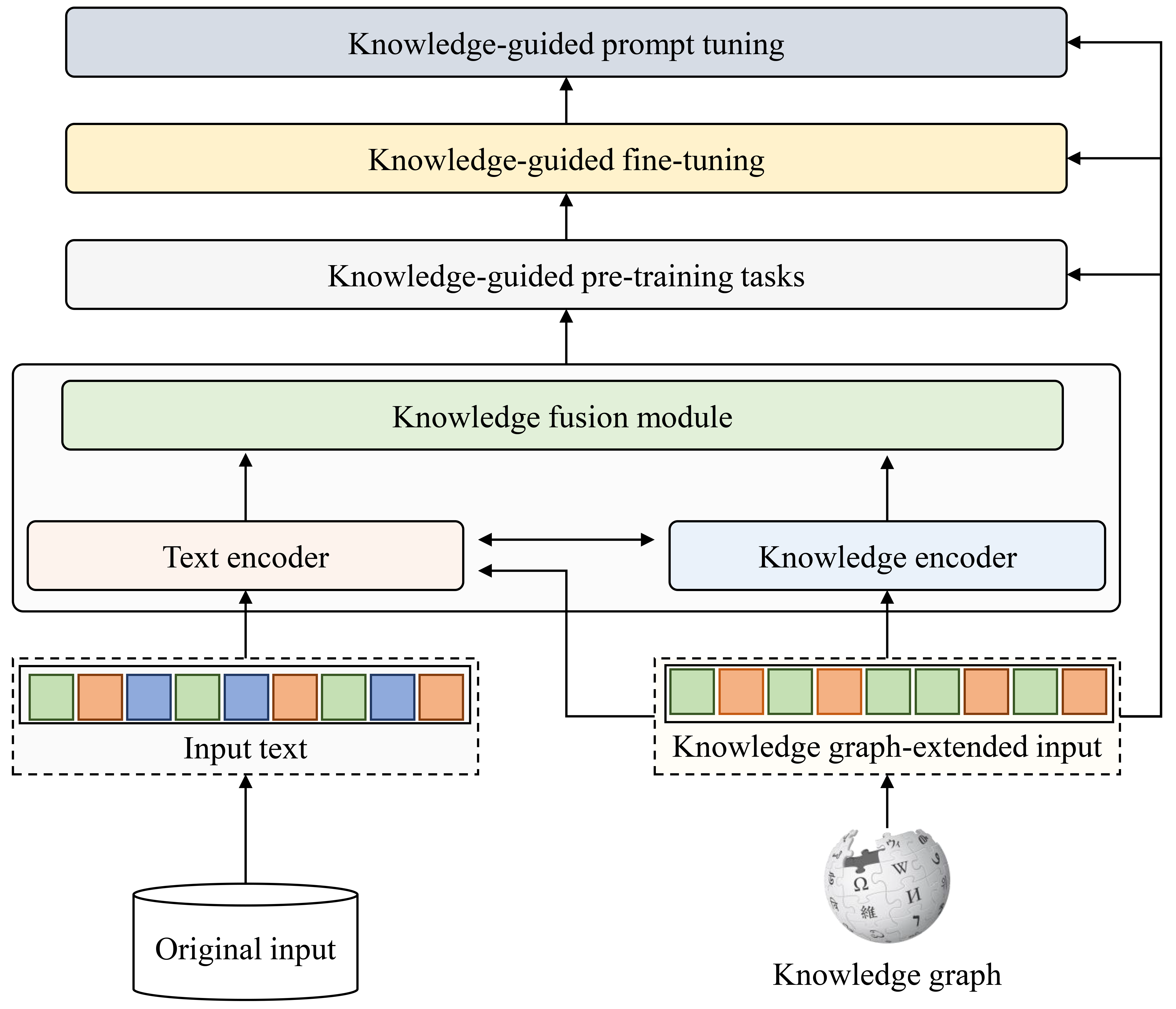}
    \caption{Technical framework of developing KGLLMs.}
    \label{fig:kgllm}
\end{figure}

The development framework for KGLLMs based on existing technologies is depicted in Fig.~\ref{fig:kgllm}. Since LLMs primarily scale the size of parameters and training data from PLMs, their model architecture and training methods remain largely unchanged. Hence, all three types of KGPLM methods introduced before can be applied to developing KGLLMs. The before-training enhancement approaches can be utilized to construct KG-extended text, improving input quality and integrating factual information into the input. The during-training enhancement methods can be employed to adaptively fuse textual knowledge and structural knowledge to learn knowledge-enhanced word representations. Graph encoders, such as GNN, can serve as knowledge encoders, while attention mechanisms can be utilized to design the knowledge fusion module. Multi-task learning, including knowledge-guided pre-training tasks, helps improve LLMs' learning of factual knowledge. The post-training enhancement methods can be utilized to further improve the performance of LLMs on some domain-specific tasks by fine-tuning them on knowledge-extended data or knowledge-grounded tasks. Moreover, one of the most important recent advancements of LLMs is prompt learning, which effectively improves the quality of generated text and enhance LLMs' generalization capability by inserting text pieces into the input. In prompt learning, selecting suitable prompt templates for specific tasks is crucial for enhancing model performance, requiring domain expertise. Therefore, KGs can be integrated into constructing prompt templates to make use of domain knowledge, which is expected to improve the model's understanding of domain factual knowledge by guiding LLMs with knowledge prompts.

\subsection{Discussion and Future Directions}


In addition to knowledge graph enhancement methods, there are also other enhancement methods that can be used to improve LLMs' factual language modeling ability. Typically, these methods include data augmentation and retrieval augmentation. Data augmentation involves refining the training data during pretraining and emphasizing informative words, emphasizing the importance of the training corpus in equipping the model with factual knowledge. Compared with knowledge graph enhancement methods, these approaches utilize implicit knowledge to model factual knowledge in text and ignore the relationships between entities. Retrieval augmentation has emerged as a widely adopted approach, allowing LLMs to retrieve external data from databases \cite{pmlr-v162-borgeaud22a} or tools and pass it to LLMs in the form of prompts or embeddings to improve LLMs' generations. These methods can address some challenges faced by plain LLMs, such as outdated information and the inability to memorize. However, they cannot fundamentally improve LLMs' knowledge modeling ability since they do not change LLMs' parameters.

Besides, some plugins have been developed to enhance the capabilities of LLMs in the context of a knowledge base. For example, the Browsing plugin can call search engines to access real-time information on the website; the Retrieval plugin\footnote{https://github.com/openai/chatgpt-retrieval-plugin} uses OpenAI embeddings to index and search documents in vector databases; the Wolfram\footnote{https://www.wolfram.com/wolfram-plugin-chatgpt/} plugin enables ChatGPT to provide more comprehensive and accurate answers by giving it access to the Wolfram Alpha knowledge base; the Expedia plugin\footnote{https://chatonai.org/expedia-chatgpt-plugin} enables ChatGPT to provide personalized travel recommendations with the help of Expedia's entity graph.

Although KGLLMs have achieved some success, there are still many unresolved challenges. Here, we outline and discuss a few promising research directions for KGLLMs.


\textbf{Improving the efficiency of KGLLMs.} Due to the need for preprocessing and encoding knowledge from KGs, developing KGLLMs typically requires more computational resources and time compared to plain LLMs. However, the scaling law of KGLLMs may differ from that of plain LLMs. Previous studies on KGPLMs have demonstrated that smaller KGPLMs can even outperform larger PLMs. Therefore, a comprehensive investigation of the scaling law of KGLLMs is necessary to determine the optimal parameter size for their development. Based on this, we can potentially achieve a smaller model that satisfies performance requirements, resulting in reduced computational resources and time.


\textbf{Merging different knowledge in different ways.} Some common and well-defined knowledge could be stored within KGs for ease of access, while rarely used or implicit knowledge that cannot be expressed through triples should be incorporated into the parameters of LLMs. In particular, domain-specific knowledge, although infrequently accessed, may still require a significant amount of human effort to construct an associated KG due to the sparse nature of its related corpus.


\textbf{Incorporating more types of knowledge.} As introduced in Section~\ref{sec:3}, the majority of existing KGPLMs only utilize a single modality and static KGs. However, there exist multimodal and temporal KGs that contain multimodal and temporal knowledge. These types of knowledge can complement textual and structural knowledge, enabling LLMs to learn the relationships between entities over time. Moreover, multimodal pre-trained models have gained popularity as they have been proven to improve the performance of pre-trained models on multimodal tasks \cite{PALM} and enhance their cognitive ability. Therefore, incorporating multimodal and temporal KGs into LLMs has the potential to improve their performance, which is worth investigating. To achieve this goal, we need to align multimodal entities, design encoders capable of processing and fusing multimodal temporal data, and establish multimodal temporal learning tasks to extract useful information.


\textbf{Improving the effectiveness of knowledge incorporation.} By modifying inputs, model architecture, and the fine-tuning process, diverse methods have been proposed to incorporate relational triplets into PLMs. However, each method has its own set of advantages and disadvantages, with some performing well on particular tasks but underperforming on others. For example, LUKE \cite{LUKE} exhibits superior performance over KEPLER \cite{KEPLER} in most entity typing and relation classification tasks but performs worse in a few other tasks \cite{CokeBERT}. Besides, recent experimental analysis \cite{hou2022understanding} reveals that existing KGPLMs integrate only a small fraction of factual knowledge. Therefore, there is still a lot of room for research on effective knowledge integration methods. Further research is required on the selection of valuable knowledge and avoiding catastrophic forgetting when faced with vast and clashing knowledge.



\textbf{Enhancing the interpretability of KGLLMs.} Although it is widely believed that KGs can enhance the interpretability of LLMs, corresponding methods have not yet been thoroughly studied. Schuff \textit{et al.} \cite{schuff-etal-2021-external} investigated whether integrating external knowledge can improve natural language inference models' explainability by evaluating the scores of generated explanations on in-domain data and special transfer datasets. However, they found that the most commonly used metrics do not consistently align with human evaluations concerning the accuracy of explanations, incorporation of common knowledge, and grammatical and labeling correctness. To provide human-understandable explanations for LLMs, Chen \textit{et al.} \cite{LMExplainer} proposed a knowledge-enhanced interpretation module that utilizes a KG and a GNN to extract key decision signals of LLMs. Despite a few studies attempting to improve the interpretability of PLMs, it remains unclear how to leverage KGs to improve the interpretability of KGPLMs. A feasible approach may involve searching for the relevant reasoning path in KGs based on the generated content and then generating an explanatory text based on the reasoning path.


\textbf{Exploring domain-specific KGLLMs.} Though there is already considerable research incorporating standard KGs with general PLMs, limited work has focused on domain-specific KGLLMs. However, the rise of artificial intelligence for science will lead to an increasing demand for domain-specific KGLLMs. In comparison to general LLMs, domain-specific LLMs require greater precision and specificity in incorporating domain knowledge. As a result, constructing accurate domain-specific KGs and integrating them with LLMs warrant further exploration. In order to develop domain-specific KGLLMs, it is essential to first construct a domain KG and gather relevant corpus data with the help of domain experts. Considering the generality of language patterns, it is advisable to blend common KGs with the domain-specific KG for enhancement.

\section{Conclusion}

The phenomenal success of ChatGPT has spurred the rapid advancement of LLMs. Given the impressive performance of LLMs on a variety of NLP tasks, some researchers wonder if they can be viewed as a type of parameterized knowledge base and replace KGs. However, LLMs still fall short in recalling and correctly using factual knowledge while generating knowledge-grounded text. In order to clarify the value of KGs in the era of LLMs, a comprehensive survey on KGPLMs was conducted in this paper. We began by examining the background of PLMs and the motivation for incorporating KGs into PLMs. Next, we categorized existing KGPLMs into three categories and provided details about each category. We also reviewed the applications of KGPLMs. After that, we analyzed whether PLMs and recent LLMs can replace KGs based on existing studies. In the end, we proposed enhancing LLMs with KGs to conduct fact-aware language modeling for improving their learning of factual knowledge. This paper addresses three questions: (1) What is the value of KGs in the era of LLMs? (2) How to incorporate KGs into LLMs to improve their performance? (3) What do we need to do for the future development of KGLLM? We hope this work will stimulate additional research advancements in LLM and KG.

\bibliographystyle{IEEEtran}
\bibliography{reference}

\vfill

\end{document}